%% file: main.tex
\documentclass{article} % For LaTeX2e
\usepackage[table]{xcolor}
\usepackage[final]{colm2026_conference}

\usepackage{microtype}
\usepackage{hyperref}
\usepackage{url}
\usepackage{booktabs}
\usepackage{amsmath}
\usepackage{subfiles}
\usepackage{graphicx}
\usepackage{multirow}
\usepackage{fancyvrb}
\usepackage[most]{tcolorbox}

\usepackage{lineno}
\usepackage{xspace}

\usepackage{enumitem}

\newcommand\benchmark{\textsc{MedConceal}\xspace}

\definecolor{darkblue}{rgb}{0, 0, 0.5}
\definecolor{fairbest}{HTML}{D6EAF8}
\definecolor{overallbest}{HTML}{D5F5E3}
\hypersetup{colorlinks=true, citecolor=darkblue, linkcolor=darkblue, urlcolor=darkblue}

\graphicspath{{images/}}

\title{MedConceal: A Benchmark for Clinical Hidden-Concern Reasoning Under Partial Observability}

\author{
Yikun Han\textsuperscript{1}, Joey Chan\textsuperscript{1}, Jingyuan Chen\textsuperscript{1}, Mengting Ai\textsuperscript{1}, Simo Du\textsuperscript{2}, Yue Guo\textsuperscript{1}\\
\textsuperscript{1}University of Illinois Urbana-Champaign\\
\textsuperscript{2}NYC Health + Hospitals/Jacobi\\
\texttt{\{yikunh2, yueg\}@illinois.edu}
\quad
\href{https://github.com/FAIRHealth/MedConceal}{%
\raisebox{-0.22ex}{\includegraphics[height=1.05em]{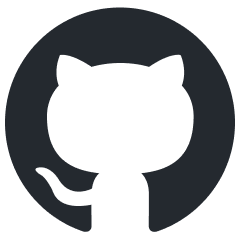}}\,\texttt{GitHub}}
\quad
\href{https://huggingface.co/datasets/Eclipse42/MedConceal}{%
\raisebox{-0.28ex}{\includegraphics[height=1.12em]{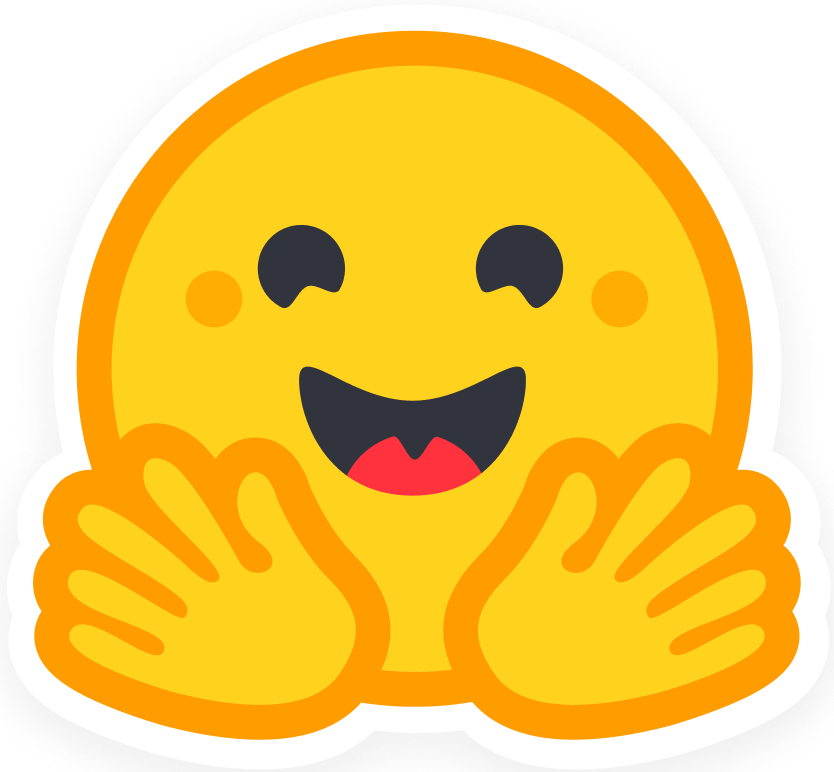}}\,\texttt{Hugging Face Dataset}}
}
\newtcolorbox{promptbox}{
  breakable,
  colback=gray!3,
  colframe=gray!50,
  boxrule=0.4pt,
  arc=1mm,
  left=1mm,
  right=1mm,
  top=1mm,
  bottom=1mm
}

\begin{document}

\ifcolmsubmission
\linenumbers
\fi

\maketitle

\begin{abstract}
Patient-clinician communication is an asymmetric-information problem: patients often do not disclose fears, misconceptions, or practical barriers unless clinicians elicit them skillfully. 
Effective medical dialogue therefore requires reasoning under partial observability: clinicians must elicit latent concerns, confirm them through interaction, and respond in ways that guide patients toward appropriate care.
However, existing medical dialogue benchmarks largely sidestep this challenge by exposing hidden patient state, collapsing elicitation into extraction, or evaluating responses without modeling what remains hidden. 
We present \benchmark, a benchmark with an interactive patient simulator for evaluating hidden-concern reasoning in medical dialogue, comprising 300 curated cases and 600 clinician-LLM interactions.
Built from clinician-answered online health discussions, each case pairs clinician-visible context with simulator-internal hidden concerns derived from prior literature and structured using an expert-developed taxonomy. 
The simulator withholds these concerns from the dialogue agent, tracks whether they have been revealed and addressed via theory-grounded turn-level communication signals, and is clinician-reviewed for clinical plausibility. This enables process-aware evaluation of both task success and the interaction process that leads to it.
We study two abilities: \emph{confirmation}, surfacing hidden concerns through multi-turn dialogue, and \emph{intervention}, addressing the primary concern and guiding the patient toward a target plan. 
Results show that no single system dominates: frontier models lead on different confirmation metrics, while human clinicians (N=159) remain strongest on intervention success. Together, these results identify hidden-concern reasoning under partial observability as a key unresolved challenge for medical dialogue systems.
\end{abstract}

\input{sections/01_intro.tex}

\input{sections/02_framework_overview}

\input{sections/03_experiments}

\input{sections/04_results}

\input{sections/06_conclusion}

\section*{Ethics Statement}
This work uses two human-data sources: public clinician-responded threads from the Reddit forum \textit{r/AskDocs}, and a controlled Prolific study with clinician participants. Because the benchmark is derived from real-world health communication, we considered privacy, consent, representation, and downstream misuse throughout case construction and evaluation. Benchmark cases are converted into de-identified structured profiles with clinician-visible chart fields and simulator-internal hidden concerns, and case construction is grounded in an expert-developed taxonomy with physician plausibility review. The clinician-participant study received IRB approval, and all participants provided informed consent before participation. Participants were permitted to stop at any point during consent, the demographic survey, or the assigned study tasks, and compensation was prorated to reflect the portion of the study they had already completed. We report participant demographics only in aggregate and exclude directly identifying information from the benchmark artifacts. Our experiments use a mix of publicly available models and commercial APIs, but \benchmark is intended as an evaluation benchmark rather than a clinical deployment system. We therefore emphasize risks such as hallucinated medical advice, unsupported inferences about patients, benchmark gaming, and the reinforcement of demographic or communication biases, and use the benchmark to expose these failure modes rather than to endorse autonomous clinical use.

\section*{Reproducibility Statement}
We provide the benchmark code, configuration, and analysis artifacts needed to reproduce the main construction and evaluation pipeline.\footnote{Anonymous repository: \url{https://anonymous.4open.science/r/MedConceal}.} 

\bibliography{ref}
\bibliographystyle{colm2026_conference}

\appendix
\section{Appendix Table of Contents}
\begin{itemize}
    \item \textbf{Appendix B: Related Work.}
    \item \textbf{Appendix C: Data and Case Construction}: source data and conversation extraction; common-condition subset construction; patient profile taxonomy; patient case population; annotation and labeling.
    \item \textbf{Appendix D: Human Clinician Study Details}: recruitment and eligibility criteria; study procedure and interface; study platform screenshots; instructions and materials; survey instrument; demographics; average dialogue lengths.
    \item \textbf{Appendix E: Evaluation Details}: concern matching and statistical comparisons; intervention success gate; process and style metrics; evaluator validation and state-tracking sensitivity; concern-type and severity analyses; human-reference coverage; scope, safety, and external-validity limitations.
    \item \textbf{Appendix F: Implementation Details}: patient simulator; turn evaluator and latent policy; rubric feature vector.
    \item \textbf{Appendix G: Usage of LLMs.}
\end{itemize}
\input{sections/05_related_work}

\input{sections/07_appendix}

\end{document}

%% file: sections/01_intro.tex
\section{Introduction}

Interactive language models have achieved impressive performance in many settings \citep{singh2025openai, guo2025deepseek, qwen3.5}, but remain challenged by reasoning in asymmetric-information interactions: where critical information is private, costly to elicit, or strategically withheld \citep{peng2025communication, xia2024measuring}. Patient--clinician communication is a canonical example: patients often do not volunteer key concerns or contextual constraints (e.g., embarrassment, stigma, fear of judgment), which can undermine shared understanding and shared decision-making and is associated with worse treatment adherence \citep{levy2018prevalence}. Any benchmark for this setting must preserve this asymmetry: hidden concerns should remain private and become accessible only after clinically appropriate, empathic elicitation.

Yet many medical dialogue benchmarks implicitly assume a fully forthcoming patient: either (a) the model receives the complete patient profile, including psychosocial context and concerns \citep{fan2025ai}, or (b) success is scored primarily on extracting facts that a patient would readily disclose even without careful elicitation \citep{li2024mediq,lialfa}. These setups conflate elicitation with extraction and under-evaluate what makes clinical communication hard: building rapport, asking targeted open-ended follow-ups, and uncovering psychosocial barriers that shape downstream decisions.

However, there are three main challenges in building and evaluating benchmarks for hidden-concern communication. First, hidden concerns are rarely annotated in natural data, so we must construct cases with realistic surface narratives and well-defined held-out concerns without turning the task into a scripted checklist. Second, the patient simulator must be stateful and controllable: it should separate clinician-visible chart context from simulator-internal psychosocial information, resist prompt leakage and meta-gaming, and reveal concerns only when elicitation signals are strong. Third, evaluation must be process-aware: it should distinguish whether a concern was actually elicited (and when), whether it was addressed, and whether the interaction supported the intended plan: without rewarding vague inferences unsupported by the dialogue.

We convert clinician-responded online health discussions into standardized patient cases and use them to benchmark two linked abilities. In \emph{confirmation}, a clinician must uncover multiple hidden concerns through natural dialogue. In \emph{intervention}, the clinician must address the most salient concern and guide the patient toward a target plan (e.g., further testing, treatment uptake) while maintaining empathy and safety-netting. Our framework couples a reserved patient simulator with theory-grounded conversational signals so that progress depends on what the clinician says, not on privileged access to hidden state.\footnote{The study platform used for the human-clinician data collection is available at \url{https://frontend-prolific.up.railway.app/}.}

Our main contributions are:
\begin{itemize}
\item A benchmark of patient--clinician conversations with held-out hidden concerns, spanning both confirmation (elicitation) and intervention (addressing and persuasion) tasks.
\item A reserved, stateful patient simulator that withholds concerns by default and reveals them only under clinically meaningful elicitation signals, enabling realistic asymmetric-information interaction.
\item An evaluation protocol that separates elicitation and extraction, uses strict concern matching to keep precision/recall interpretable, and supports comparisons between human clinicians and LLM-based clinician agents.
\end{itemize}

%% file: sections/02_framework_overview.tex
\section{Framework Overview}
\label{sec:framework_overview}

This section summarizes case construction, the partially observed interaction setting, and the patient-agent dynamics used for confirmation and intervention.

\subsection{Patient Simulator Construction}
Benchmark cases are derived from real-world clinician-responded conversations on the Reddit forum \textit{r/AskDocs} and converted into structured patient profiles with clinician-visible chart fields, simulator-internal psychosocial fields, and intervention metadata defined by a clinician-developed taxonomy (details in Appendix~\ref{patient_taxonomy}). From the extracted pool, we constructed a common-condition subset so that clinicians could focus on confirmation and intervention quality rather than being disproportionately penalized by very rare or highly specialized presentations.
Hidden concerns are organized according to a four-category taxonomy, refined in collaboration with clinicians, including:
\begin{itemize}[noitemsep, topsep=1pt, leftmargin=10pt]
    \item \textbf{Misinformation or Misconceptions}: inaccurate beliefs about diagnosis, treatment, risk, side effects, or expected outcomes that shape care decisions, including barrier-forming medication beliefs and mistrust-driven interpretations of medical recommendations \citep{horne2013understanding,cuevas2019mistrust}.
    \item \textbf{Emotional Discomfort or Fear}: hesitation or resistance driven by fear, anxiety, shame, or stigma, consistent with work on patients' emotional cues and concerns in medical consultations \citep{zimmermann2011vrcodes,del2017verona}.
    \item \textbf{Communication Barriers}: difficulty disclosing concerns or understanding recommendations because of explanation mismatch, health literacy limitations, language discordance, or limited communicative alignment \citep{williams2002healthliteracy,alshamsi2020languagebarriers}.
    \item \textbf{Financial or Insurance-Related Concern}: barriers related to cost, insurance coverage, transportation, or work disruption, reflecting literature on affordability discussions and cost-related nonadherence \citep{wilson2007physician,patel2014cost}.
\end{itemize}

Before deployment, we iteratively reviewed pilot cases and simulator behavior with clinician collaborators to ensure that the roleplay remained clinically plausible rather than simply reproducing source-thread text. The final simulator was evaluated by four clinician-guided criteria:
\begin{itemize}[noitemsep, topsep=1pt, leftmargin=10pt]
    \item \textbf{Calibrated patient turn structure.} Replies stay short, turn-responsive, and introduce only one or two issues at a time.
    \item \textbf{Health-literacy realism.} Wording is aligned to the patient's background and avoids unintroduced jargon.
    \item \textbf{Progressive disclosure.} Hidden concerns and background facts are revealed gradually rather than volunteered all at once.
    \item \textbf{Non-trivial intervention dynamics.} Intervention cases require concern-specific addressing before acceptance and allow realistic pushback and clarification.
\end{itemize}
Case sourcing is detailed in Appendix~\ref{sec:appendix_patient_population}, pilot-study and human-study design in Appendix~\ref{sec:appendix_human_clinician}.

\subsection{Interaction Setting and Observability}
\begin{figure}[t]
    \centering
    \includegraphics[width=1\linewidth]{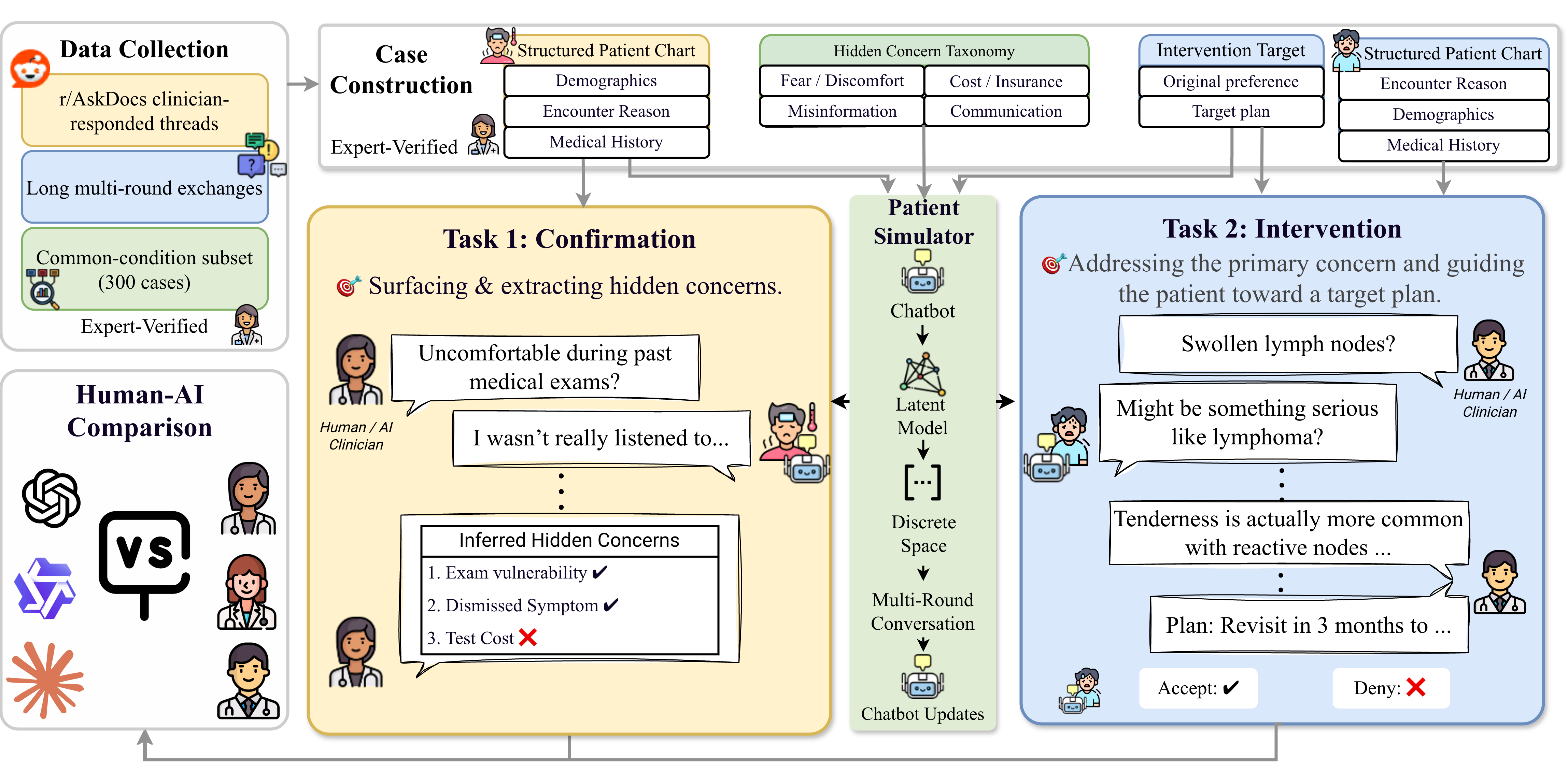}
    \caption{Framework overview. Cases are collected from real-world clinician-responded \textit{r/AskDocs} conversations, filtered to longer common-condition exchanges, and converted into structured patient charts, intervention targets, and hidden-concern annotations with clinician-guided verification. The patient simulator exposes only clinician-visible information while keeping hidden concerns internal. In Task~1 (confirmation), the clinician must surface and extract fine-grained hidden concerns through natural dialogue rather than meta-category probing. In Task~2 (intervention), the clinician must address the primary concern and guide the patient toward the target plan. The same framework is used to compare AI clinician agents and human clinicians under matched observable information.}
\label{fig:framework_overview}
\end{figure}

As shown in Figure~\ref{fig:framework_overview}, each case defines (i) \emph{clinician-visible} chart context and (ii) \emph{simulator-internal} psychosocial state with hidden concerns that are not directly observable.

The clinician (human or LLM agent) interacts with the patient simulator via natural dialogue; hidden concerns should become accessible only after clinically appropriate elicitation. We evaluate two abilities: \emph{confirmation} (surfacing and extracting hidden concerns) and \emph{intervention} (addressing the primary concern and guiding the patient toward a target plan). The same clinician-visible chart is shown to human clinicians and passed to AI clinician agents, while hidden concerns remain hidden from both.

\subsection{Patient Agent Dynamics}

We model the patient agent as a discrete-time dynamical system with (i) a discrete concern state and (ii) a continuous latent evidence state over hidden concerns $\{c_1, c_2, \dots, c_k\}$.

\subsubsection{State Space Representation}
At each conversation turn $t$, the status of the $i$-th concern is tracked by a state variable $s_{t,i} \in \{0, 1, 2\}$, defined as:
\[
s_{t,i} = 
\begin{cases} 
0 & \text{Hidden} \\
1 & \text{Revealed} \\
2 & \text{Addressed}
\end{cases}
\]
The patient state at turn $t$ is
\begin{equation}
\mathbf{S}_t = [s_{t,1}, s_{t,2}, \dots, s_{t,k}]^\top \in \{0,1,2\}^k.
\end{equation}

We also maintain continuous latent evidence. $\mathbf{E}_t$ stores per-concern reveal evidence, while intervention additionally tracks a scalar addressing evidence $A_t$ for the single targeted concern.

\subsubsection{Theory-Grounded Signal Backbone}
We extract turn-level rubric signals from established communication frameworks: (i) \textbf{RIAS} \citep{roter2002roter} for interaction function and socio-emotional behavior, (ii) \textbf{VR-CoDES} \citep{del2017verona} for concern cue elicitation and clinician response space, and (iii) \textbf{Necessity-Concerns Framework (NCF)} \citep{horne2013understanding} for adherence-oriented persuasion quality. We formulated these into a 10-dimensional rubric feature vector $z_t \in [0,1]^{10}$ scored per clinician turn, comprising data gathering, emotional responsiveness, partnership/activation, concern elicitation, space provision, necessity support, concern mitigation, plan specificity/safety-netting, pending-question coverage, and meta-probe risk. Appendix~\ref{sec:appendix_turn_evaluator} and Table~\ref{tab:rubric_features} give the implementation details and feature definitions.

\subsubsection{Task 1: Confirmation}
Let $u_t$ be the clinician utterance at turn $t$ and let $H_t$ be the recent dialogue-history window used by the shared turn evaluator. The evaluator maps $(u_t, H_t)$ to the rubric feature vector $z_t=\big(r_{t,d}\big)_{d\in\mathcal{D}} \in [0,1]^{10}$, where $r_{t,d}$ is the score for rubric dimension $d$ and $\mathcal{D}$ indexes the 10 rubric dimensions in Table~\ref{tab:rubric_features}. We also compute a concern-target overlap score $o_{t,i} \in [0,1]$ from lexical overlap between $u_t$ and concern $c_i$.  

\textbf{Reveal observation and evidence update.}
For each concern $c_i$, we compute a reveal observation probability using a learned logistic model:

\begin{equation}
p^{\text{reveal}}_{t,i} = \sigma\!\left((w+\delta_{\kappa(i)})^\top \phi(z_t, o_{t,i})\right).
\end{equation}
Here $\sigma$ is the sigmoid, $\phi(z_t,o_{t,i})=[z_t;o_{t,i}]$, $w$ is a shared weight vector, and $\delta_{\kappa(i)}$ is a concern-cluster offset. Reveal evidence is then updated with an exponential moving average:
\begin{equation}
E_{t+1,i} = \alpha E_{t,i} + (1-\alpha)\,p^{\text{reveal}}_{t,i}.
\end{equation}
If $u_t$ is classified as meta-category probing, the anti-gaming rule blocks reveal updates for that turn.

\textbf{Hysteresis transition.}
$s_{t,i}$ transitions only when evidence crosses a high threshold or sustains a lower threshold for multiple consecutive turns:
\begin{equation}
s_{t+1,i} =
\begin{cases}
1 & \text{if } s_{t,i}=0 \land \left(E_{t+1,i}\ge T_{\text{hi}} \;\;\lor\;\; \text{hits}(E_{t+1,i}\ge T_{\text{lo}})\ge N\right) \\
s_{t,i} & \text{otherwise}.
\end{cases}
\end{equation}

\subsubsection{Task 2: Intervention}
Once a concern is revealed ($s_{t,i}=1$), the clinician must address that concern and move the patient toward the target plan. We compute an addressing observation probability $p^{\text{addr}}_t$ from the same rubric feature vector $z_t$, then update addressing evidence with an EMA:
\begin{equation}
A_{t+1} = \beta A_t + (1-\beta)\,p^{\text{addr}}_t.
\end{equation}
Intervention transitions use a stricter gate than confirmation. Let $\tau_i$ denote the first turn at which concern $c_i$ becomes revealed. A turn is eligible only if $s_{t,i}=1$ and $(t-\tau_i)\ge L$. We define an addressing hit when both the per-turn quality gate and the EMA gate are satisfied:
\begin{equation}
h_t = \mathbb{I}\!\left(p^{\text{addr}}_t \ge \eta \;\land\; A_{t+1}\ge T^{A}\right),
\end{equation}
and require $K$ consecutive eligible hits before transitioning to addressed:
\begin{equation}
s_{t+1, i} = 
\begin{cases} 
2 & \text{if } s_{t,i} = 1 \land (t-\tau_i)\ge L \land \text{hits}(h_t=1)\ge K \\
s_{t,i} & \text{otherwise}
\end{cases}
\end{equation}

\textbf{Intervention gate.}
The intervention gate uses the primary concern $c^*$:
\begin{equation}
\mathbb{I}_{\text{gate}}(\mathbf{S}_t) = \mathbb{I}(s_{t, c^*} = 2)
\end{equation}

\textbf{Post-gate patient continuation.}
The intervention endpoint used in this paper is whether the primary concern ever reaches the addressed state by the end of the dialogue, i.e., whether there exists any turn $t \le T$ such that $s_{t,c^*}=2$.

%% file: sections/03_experiments.tex
\section{Experiments}

\textbf{Tasks and Interaction Protocol}
We evaluate two reserved-patient tasks. In \textbf{Task 1 (confirmation)}, the clinician receives clinician-visible chart context, conducts a dialogue to elicit hidden psychosocial concerns, and then submits structured findings (concern categories plus short descriptions). In \textbf{Task 2 (intervention)}, the clinician is assigned a separate case and conducts a new dialogue to address the primary concern and guide the patient toward a target plan. Humans and AI systems receive the same visible case summary, comprising the patient description, encounter reason, and medical/surgical history. Intervention encounters additionally display the patient's initial preference and target plan (Appendix Figures~\ref{fig:platform_confirmation_chart} and~\ref{fig:platform_intervention_page}). Holding this observable context fixed isolates the communication problem of eliciting and addressing information that is not initially available.
Patient population details (case sources, inclusion criteria, and summary statistics) are reported in Appendix~\ref{sec:appendix_patient_population}.

\textbf{Human Clinicians}
We recruited 159 clinician participants through Prolific because it supports role-based screening and has been reported to provide strong online data quality \citep{peer2022data}. The clinician pool was relatively young (mean age 31.4, median 31, range 20--71), 68.1\% male and 31.0\% female, and concentrated in general medical doctors/non-specialists (41.6\%), family physicians/general practitioners (15.0\%), other specialist doctors (13.3\%), and emergency medicine specialists (10.6\%); 64.9\% reported 1--5 years of experience. We filtered on healthcare job roles and English fluency; full eligibility, remaining demographic breakdowns, survey items, and interface details are reported in Appendix~\ref{sec:appendix_human_clinician}.

The study received IRB approval. Before full deployment, pilot runs with a clinician collaborator and a small Prolific sample were used to refine workflow, wording, and time limits. To avoid cross-task information leakage, no clinician was assigned the same patient case in both the confirmation and intervention conditions.

\textbf{AI Clinicians}
We evaluate five AI clinician baselines under the same benchmark setup: Qwen-3.5-9B \citep{qwen3.5}, Claude Sonnet 4.5, GPT-5.2 \citep{singh2025openai}, Doctor-R1 \citep{lai2025doctor}, and Llama3-OpenBioLLM-8B \citep{openbiollm8b}. In the main results, each model is reported under a fixed 8-turn setting, chosen to approximately match the mean human dialogue length (8.2 turns in both tasks), and a longer 20-turn setting. In confirmation, the longer setting allows STOP/CONTINUE decisions after turn 5 with a hard cap of 20 turns. In intervention, adaptive self-stop is disabled; runs terminate only when the primary concern is marked addressed or when the conversation reaches the 20-turn cap.

\textbf{Evaluation Metrics}
\label{sec:evaluation_metrics}
We report both \emph{performance} metrics and complementary \emph{communication-style} metrics. Performance metrics combine patient-state traces with post-dialogue finding matching; the trace-derived quantities depend on the \texttt{evaluator\_analysis} fields emitted by the shared turn evaluator in Section~\ref{sec:framework_overview}, while the submitted-finding metrics are computed from the final dialogue transcript and gold concerns (Appendix~\ref{sec:appendix_concern_matching}). Communication-style metrics are computed from transcript-level analyses, including a post-hoc style-analysis pipeline and question-structure summaries (Appendix~\ref{sec:appendix_style_metrics}). All metrics are computed in the same way across humans and AI clinicians, and the human condition is treated as a reference baseline rather than a gold standard.

\begin{itemize}[noitemsep, topsep=1pt, leftmargin=10pt]
\item \textbf{Performance metrics.} These metrics summarize what the clinician elicited and whether the key barrier was eventually addressed.
\item \textbf{Confirmation.} We report \emph{Reveal Rate}, the fraction of gold concerns surfaced during the dialogue, because it isolates whether concerns were actually elicited before any post-dialogue summarization. After the dialogue, the clinician submits structured findings (category plus short description); AI systems produce the same artifact with a post-dialogue extraction prompt. We score these findings at two levels: coarse-grained Precision/Recall/F1 over the four benchmark categories (\emph{Misinformation or Misconceptions}, \emph{Emotional Discomfort or Fear}, \emph{Communication Barriers}, and \emph{Financial or Insurance-Related Concern}) to test whether the correct \emph{type} of barrier was recovered, and fine-grained Precision/Recall/F1 over case-specific concern statements to test whether the patient's actual concern was recovered. We additionally report \emph{matched-but-no-reveal} (MBNR) to distinguish grounded elicitation from correctly formatted guessing.
\item \textbf{Intervention.} Each case designates a primary concern. We report \emph{Success}, because the main intervention question is whether the clinician ultimately addresses the key barrier. We also report \emph{Reveal Rate} to show whether the hidden state was surfaced before intervention, \emph{Turn-to-Address} and \emph{Reveal-to-Address} to separate slow elicitation from slow response, and \emph{Meta-Probe}, which flags turns that probe the benchmark categories directly instead of through natural questioning.
\item \textbf{Communication-style metrics.} These metrics characterize how the dialogue is conducted rather than whether the simulator objective is achieved. For confirmation, we report Empathy, Collaboration, Problem-Solving, Values Det., Early Open, Readability, and Word/Turn to capture relational quality, probing strategy, value sensitivity, readability, and verbosity. For intervention, we report Empathy, Rationale, Problem-Solving, Actionability, Early Open, Readability, and Word/Turn to characterize explanation quality, probing strategy, and actionability.
\end{itemize}

%% file: sections/04_results.tex
\section{Results}
We report both matched-length 8-turn results and extended 20-turn results for humans and AI clinicians.
At the matched 8-turn setting, human clinicians remain strongest on the most specific performance metrics, especially fine-grained confirmation recovery and intervention success. For confirmation, the adaptive runs are materially longer than the observed human conversations, averaging 11.7--19.4 turns depending on the model. We therefore treat the 20-turn setting as a separate extended-turn condition rather than a direct matched-length comparison, since longer AI runs can improve some results but typically do so with later addressing and longer dialogues.

\subsection{Task 1: Confirmation}
\label{sec:results_confirmation}
\noindent The confirmation stage tests whether the clinician can surface hidden concerns through dialogue. We therefore focus on elicitation quality: whether patients disclose those concerns during the conversation and whether the clinician's submitted findings recover them afterward.

\textbf{Performance Metrics}
At the matched 8-turn setting, human dialogue remains strongest on grounded elicitation, while no single AI baseline dominates post-dialogue concern recovery. Table~\ref{tab:level1_human_ai} shows that human clinicians achieve the highest matched-length Reveal Rate (0.558), exceeding the best AI 8-turn results from Claude Sonnet 4.5 (0.522) and Llama3-OpenBioLLM-8B (0.520). 
On the same elicitation traces, replacing the human submitted findings with a Claude-generated sumary shifts Fine P/R/F1 from 0.721/0.411/0.523 to 0.504/0.646/0.567. Among AI 8-turn baselines, Qwen-3.5-9B gives the strongest Fine P (0.740), while OpenBioLLM achieves the strongest Coarse F1 (0.688). 
These patterns suggest that human clinicians are relatively conservative in what they submit as hidden concerns, whereas Claude-style post-dialogue summarization is more recall-oriented. Stronger elicitation during dialogue does not necessarily translate into stronger fine-grained recovery afterward. 
Under the longer adaptive setting, which is substantially longer than the human dialogues, Doctor-R1 and OpenBioLLM substantially raise Reveal Rate (0.762 and 0.944). In contrast, GPT-5.2 maintains low Reveal Rate (0.247/0.252 across settings) alongside high MBNR (0.540/0.533), indicating that its submitted findings are often weakly grounded in the dialogue and rely more on post-hoc inference than on elicited patient disclosure. Overall, no single system is best at both concern elicitation and fine-grained recovery.

\begin{table*}[t]
\centering
\small
\setlength{\tabcolsep}{2.8pt}
\renewcommand{\arraystretch}{1.08}
\resizebox{\textwidth}{!}{%
\begin{tabular}{@{}llccccccccc@{}}
\toprule
System & Setting & Reveal Rate $\uparrow$ & Coarse P $\uparrow$ & Coarse R $\uparrow$ & Coarse F1 $\uparrow$ & Fine P $\uparrow$ & Fine R $\uparrow$ & Fine F1 $\uparrow$ & MBNR $\downarrow$ \\
\midrule
\multirow{2}{*}{Human clinicians} & Human summary & \cellcolor{fairbest}0.558 & 0.791 & 0.451 & 0.574 & 0.721 & 0.411 & 0.523 & 0.227 \\
 & Claude summary & \cellcolor{fairbest}0.558 & 0.581 & \cellcolor{overallbest}0.744 & 0.652 & 0.504 & \cellcolor{overallbest}0.646 & \cellcolor{overallbest}0.567 & 0.257 \\
\midrule
\multirow{2}{*}{Qwen-3.5-9B} & 8-turn & 0.431 & 0.861 & 0.129 & 0.225 & \cellcolor{overallbest}0.740 & 0.111 & 0.193 & 0.057 \\
 & Adaptive & 0.934 & 0.452 & 0.284 & 0.349 & 0.102 & 0.064 & 0.079 & 0.006 \\
\midrule
\multirow{2}{*}{Claude Sonnet 4.5} & 8-turn & 0.522 & 0.608 & 0.631 & 0.619 & 0.494 & 0.513 & 0.503 & 0.270 \\
 & Adaptive & 0.542 & 0.573 & 0.676 & 0.620 & 0.470 & 0.555 & 0.509 & 0.297 \\
\midrule
\multirow{2}{*}{GPT-5.2} & 8-turn & 0.247 & 0.677 & 0.470 & 0.554 & 0.592 & 0.411 & 0.485 & 0.540 \\
 & Adaptive & 0.252 & 0.691 & 0.497 & 0.578 & 0.613 & 0.440 & 0.512 & 0.533 \\
\midrule
\multirow{2}{*}{Doctor-R1} & 8-turn & 0.348 & 0.731 & 0.618 & \cellcolor{fairbest}0.670 & 0.534 & 0.451 & 0.489 & 0.397 \\
 & Adaptive & 0.762 & 0.769 & 0.256 & 0.385 & 0.616 & 0.205 & 0.308 & 0.087 \\
\midrule
\multirow{2}{*}{Llama3-OpenBioLLM-8B} & 8-turn & 0.520 & \cellcolor{overallbest}0.891 & 0.114 & 0.201 & 0.633 & 0.081 & 0.143 & \cellcolor{overallbest}0.003 \\
& Adaptive & \cellcolor{overallbest}0.944 & 0.702 & 0.675 & \cellcolor{overallbest}0.688 & 0.490 & 0.471 & 0.480 & 0.013 \\
\bottomrule
\end{tabular}
}
\caption{Task 1 (confirmation) performance metrics for humans and AI clinicians. Metrics follow Section \ref{sec:evaluation_metrics}: Reveal Rate, Coarse Precision/Recall/F1, Fine Precision/Recall/F1, and MBNR (matched-but-no-reveal rate). The two human rows report the original human submitted findings and a hybrid reference that uses Claude Sonnet 4.5 summarization over the same human dialogues for the submitted-finding metrics. For each AI model, the first row uses a fixed 8-turn setting and the second row uses the adaptive confirmation policy with a hard cap of 20 turns. \colorbox{fairbest}{Blue}: best under matched-length comparison when it differs from the overall best; \colorbox{overallbest}{green}: best overall.}
\label{tab:level1_human_ai}
\end{table*}

\textbf{Communication Style Metrics}
Human clinicians remain the most concise and readable condition, while several AI baselines score higher on judge-based confirmation style dimensions such as empathy, collaboration, and problem-solving. 
In Table~\ref{tab:level1_style}, humans achieve the highest Readability (63.393) and the fewest Word/Turn (23.706). Among matched 8-turn AI systems, Claude Sonnet 4.5 scores highest on Empathy (1.900), Doctor-R1 on Collaboration (1.553), GPT-5.2 on Problem-Solving (0.990), and OpenBioLLM on Early Open score (0.482). These results suggest that judge-scored conversational style and confirmation performance are related but distinct: AI systems can score higher on perceived empathy and collaboration, while human clinicians remain more concise and readable.

\begin{table*}[t]
\centering
\small
\setlength{\tabcolsep}{3.0pt}
\renewcommand{\arraystretch}{1.08}
\resizebox{\textwidth}{!}{%
\begin{tabular}{@{}llccccccc@{}}
\toprule
System & Setting & Empathy $\uparrow$ & Collaboration $\uparrow$ & Problem-Solving $\uparrow$ & Values Det. $\uparrow$ & Early Open $\uparrow$ & Readability $\uparrow$ & Word/Turn \\
\midrule
Human clinicians & --- & 1.263 & 1.073 & 0.833 & \cellcolor{fairbest}2.663 & 0.477 & \cellcolor{overallbest}63.393 & 23.706 \\
\midrule
\multirow{2}{*}{Qwen-3.5-9B} & 8-turn & 1.487 & 1.040 & 0.447 & 1.887 & 0.279 & 59.141 & 25.054 \\
 & Adaptive & \cellcolor{overallbest}2.000 & \cellcolor{overallbest}1.760 & \cellcolor{overallbest}1.347 & \cellcolor{overallbest}3.247 & 0.458 & 53.875 & 73.161 \\
\midrule
\multirow{2}{*}{Claude Sonnet 4.5} & 8-turn & \cellcolor{fairbest}1.900 & 1.473 & 0.890 & 2.353 & 0.447 & 57.465 & 68.227 \\
 & Adaptive & 1.930 & 1.670 & 1.130 & 2.693 & 0.441 & 57.261 & 69.759 \\
\midrule
\multirow{2}{*}{GPT-5.2} & 8-turn & 1.880 & 1.543 & \cellcolor{fairbest}0.990 & 2.543 & 0.405 & 56.845 & 51.471 \\
 & Adaptive & 1.883 & 1.577 & 1.093 & 2.690 & 0.354 & 56.088 & 49.917 \\
\midrule
\multirow{2}{*}{Doctor-R1} & 8-turn & 1.737 & \cellcolor{fairbest}1.553 & 0.917 & 2.430 & 0.442 & 56.724 & 36.663 \\
 & Adaptive & 1.897 & 1.653 & 1.207 & 2.887 & 0.353 & 57.721 & 39.275 \\
\midrule
\multirow{2}{*}{Llama3-OpenBioLLM-8B} & 8-turn & 1.423 & 0.943 & 0.217 & 2.017 & \cellcolor{fairbest}0.482 & 60.564 & 42.130 \\
 & Adaptive & 1.337 & 0.870 & 0.353 & 2.243 & \cellcolor{overallbest}0.507 & 58.210 & 42.159 \\
\bottomrule
\end{tabular}
}

\caption{Task 1 (confirmation) communication-style metrics for humans and AI clinicians. Metrics follow Section \ref{sec:evaluation_metrics}: Empathy, Collaboration, Problem-Solving, Values Det. (mean number of patient values detected), Early Open, Readability, and Word/Turn. AI row ordering and color coding follow Table~\ref{tab:level1_human_ai}.}
\label{tab:level1_style}
\end{table*}

\subsection{Task 2: Intervention}
\label{sec:results_intervention}
\noindent The intervention stage tests whether the clinician can address the primary revealed concern and move the patient toward the target plan once the hidden barrier is on the table.

\textbf{Performance Metrics}
Human clinicians remain the strongest and most efficient intervention condition, while Doctor-R1 shows the largest gains with longer dialogues. Table~\ref{tab:level2_human_ai} shows that humans achieve the highest overall Reveal Rate (0.487) and tie for best Success (0.427). Doctor-R1 improves from 0.150 to 0.427 in Success and from 0.348 to 0.437 in Reveal Rate when moving from 8 to 20 turns, but its Turn-to-Address also increases from 6.938 to 11.164. These results suggest that extra turns can substantially improve intervention performance for inquiry-oriented models like Doctor-R1, but do not eliminate the human advantage in efficiency or in concern surfacing.

\begin{table*}[t]
\centering
\small
\setlength{\tabcolsep}{3.0pt}
\renewcommand{\arraystretch}{1.08}
\resizebox{\textwidth}{!}{%
\begin{tabular}{@{}llccccc@{}}
\toprule
System & Setting & Success $\uparrow$ & Reveal Rate $\uparrow$ & Turn-to-Address $\downarrow$ & Reveal-to-Address $\downarrow$ & Meta-Probe $\downarrow$ \\
\midrule
Human clinicians & --- & \cellcolor{overallbest}0.427 & \cellcolor{overallbest}0.487 & 7.125 & 3.336 & 0.018 \\
\midrule
\multirow{2}{*}{Qwen-3.5-9B} & 8-turn & 0.030 & 0.063 & 6.333 & \cellcolor{overallbest}3.000 & 0.417 \\
 & 20-turn & 0.073 & 0.073 & 8.591 & 3.045 & 0.006 \\
\midrule
\multirow{2}{*}{Claude Sonnet 4.5} & 8-turn & 0.293 & 0.313 & \cellcolor{overallbest}5.239 & 3.080 & 0.005 \\
 & 20-turn & 0.333 & 0.337 & 5.960 & 3.040 & \cellcolor{overallbest}0.004 \\
\midrule
\multirow{2}{*}{GPT-5.2} & 8-turn & 0.077 & 0.087 & 5.043 & \cellcolor{overallbest}3.000 & 0.016 \\
 & 20-turn & 0.126 & 0.140 & 8.526 & 3.053 & 0.023 \\
\midrule
\multirow{2}{*}{Doctor-R1} & 8-turn & 0.150 & 0.347 & 6.938 & \cellcolor{overallbest}3.000 & \cellcolor{overallbest}0.004 \\
 & 20-turn & \cellcolor{overallbest}0.427 & 0.437 & 11.164 & 3.039 & \cellcolor{overallbest}0.004 \\
\midrule
\multirow{2}{*}{Llama3-OpenBioLLM-8B} & 8-turn & 0.013 & 0.013 & 5.500 & \cellcolor{overallbest}3.000 & 0.013 \\
 & 20-turn & 0.040 & 0.040 & 6.417 & \cellcolor{overallbest}3.000 & 0.010 \\
\bottomrule
\end{tabular}
}
\caption{Task 2 (intervention) performance metrics for humans and AI clinicians. Metrics follow Section \ref{sec:evaluation_metrics}: Success, Reveal Rate, Turn-to-Address, Reveal-to-Address, and Meta-Probe. For every AI model, rows correspond to matched 8-turn and full 20-turn settings. Color coding follows Table~\ref{tab:level1_human_ai}.}
\label{tab:level2_human_ai}
\end{table*}

\textbf{Communication Style Metrics}
Human clinicians remain the most readable intervention condition, while GPT-5.2 scores highest on judged persuasive style metrics and OpenBioLLM most consistently asks early open-ended questions. 
Table~\ref{tab:level2_style} shows that at the matched 8-turn setting, humans achieve the highest Readability (58.831) while remaining close to the shortest AI condition in Word/Turn (30.315 vs.\ 29.484 for Qwen). GPT-5.2 leads among AI models on Empathy (1.953), Rationale (1.983), Problem-Solving (1.220), and Actionability (1.657), while OpenBioLLM achieves the highest Early Open score (0.489). Under the 20-turn setting, GPT-5.2 remains strongest on Rationale (1.993), Problem-Solving (1.377), and Actionability (1.877), and OpenBioLLM again leads on Early Open (0.542). 
These results suggest that persuasive style and early open questioning alone do not determine intervention success: despite leading on these dimensions, GPT-5.2 and OpenBioLLM do not match human performance on intervention outcome (Table~\ref{tab:level2_human_ai}).

\begin{table*}[t]
\centering
\small
\setlength{\tabcolsep}{2.2pt}
\renewcommand{\arraystretch}{1.08}
\resizebox{\textwidth}{!}{%
\begin{tabular}{@{}llccccccc@{}}
\toprule
System & Setting & Empathy $\uparrow$ & Rationale $\uparrow$ & Problem-Solving $\uparrow$ & Actionability $\uparrow$ & Early Open $\uparrow$ & Readability $\uparrow$ & Word/Turn \\
\midrule
Human clinicians & --- & 1.441 & 1.722 & 0.800 & 1.159 & 0.423 & \cellcolor{fairbest}58.831 & 30.315 \\
 \midrule
\multirow{2}{*}{Qwen-3.5-9B} & 8-turn & 1.537 & 1.550 & 0.657 & 1.030 & 0.279 & 53.901 & 29.484 \\
 & 20-turn & 1.973 & 1.947 & 1.133 & 1.837 & 0.391 & 53.070 & 76.421 \\
\midrule
\multirow{2}{*}{Claude Sonnet 4.5} & 8-turn & 1.900 & 1.883 & 0.890 & 1.207 & 0.455 & 57.465 & 68.227 \\
 & 20-turn & \cellcolor{overallbest}1.997 & 1.980 & 1.113 & 1.790 & 0.475 & \cellcolor{overallbest}62.121 & 70.100 \\
\midrule
\multirow{2}{*}{GPT-5.2} & 8-turn & \cellcolor{fairbest}1.953 & \cellcolor{fairbest}1.983 & \cellcolor{fairbest}1.220 & \cellcolor{fairbest}1.657 & 0.281 & 56.845 & 51.471 \\
 & 20-turn & 1.980 & \cellcolor{overallbest}1.993 & \cellcolor{overallbest}1.377 & \cellcolor{overallbest}1.877 & 0.263 & 49.210 & 62.904 \\
\midrule
\multirow{2}{*}{Doctor-R1} & 8-turn & 1.940 & 1.833 & 0.917 & 1.177 & 0.412 & 56.724 & 36.663 \\
 & 20-turn & 1.967 & 1.833 & 1.073 & 1.720 & 0.431 & 60.957 & 47.608 \\
\midrule
\multirow{2}{*}{Llama3-OpenBioLLM-8B} & 8-turn & 1.880 & 1.630 & 0.553 & 1.250 & \cellcolor{fairbest}0.489 & 53.642 & 53.700 \\
 & 20-turn & 1.903 & 1.637 & 0.537 & 1.290 & \cellcolor{overallbest}0.542 & 54.121 & 52.671 \\
\bottomrule
\end{tabular}
}
\caption{Task 2 (intervention) communication-style metrics for humans and AI clinicians. Metrics follow Section \ref{sec:evaluation_metrics}: Empathy, Rationale, Problem-Solving, Actionability, Early Open, Readability, and Word/Turn. Rows correspond to matched 8-turn and full 20-turn runs for every AI model. Color coding follows Table~\ref{tab:level1_human_ai}. }
\label{tab:level2_style}
\end{table*}

\begin{figure*}[t]
\centering
\includegraphics[width=\textwidth]{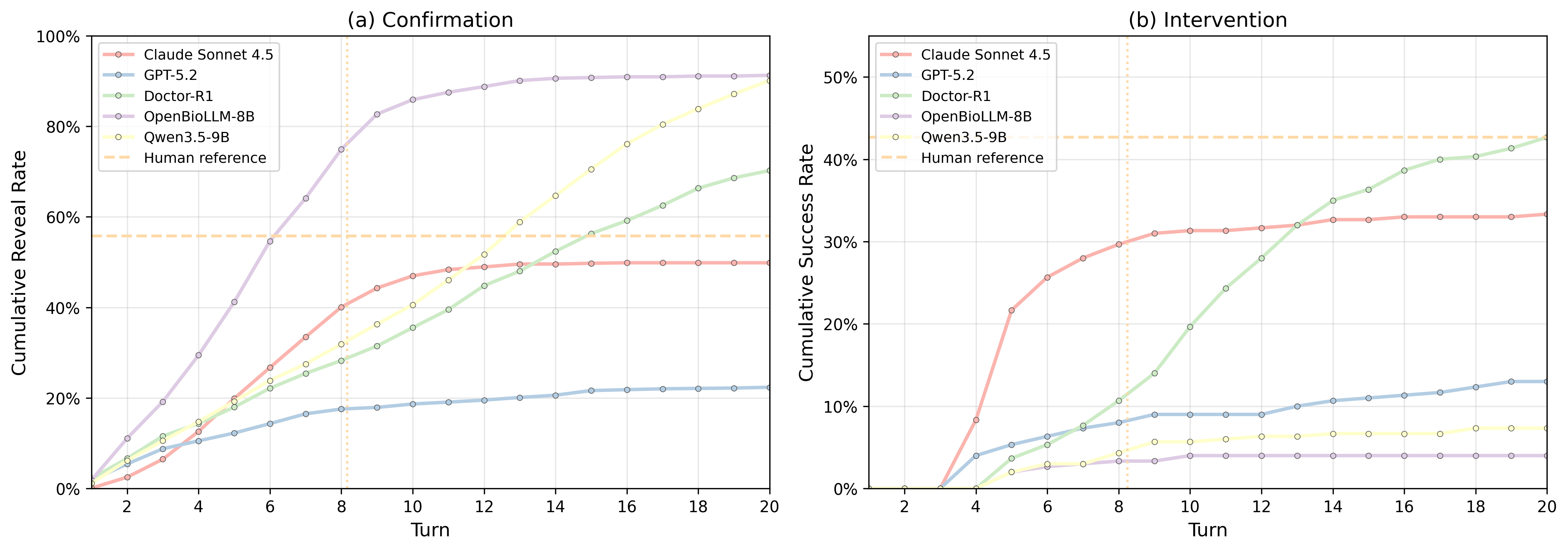}
\caption{Cumulative performance over turns for the AI clinician baselines. Panel (a) shows cumulative confirmation Reveal Rate, and panel (b) shows cumulative intervention Success. Dashed horizontal lines mark the final observed human reference level, and dotted vertical lines mark the mean observed human conversation length (8.2 turns in both tasks). Average realized dialogue lengths for all AI conditions are reported in Appendix Table~\ref{tab:avg_dialogue_lengths}.}
\label{fig:cumulative_success_by_turn}
\end{figure*}

\textbf{Performance Over Turns}
Longer dialogues benefit only a subset of AI systems rather than uniformly improving performance across both tasks. Figure~\ref{fig:cumulative_success_by_turn} shows that, in confirmation, OpenBioLLM, Qwen-3.5-9B, and Doctor-R1 continue to gain Reveal Rate well beyond the mean human dialogue length. In intervention, the clearest longer-run gains come from Doctor-R1 and Claude Sonnet 4.5, while other baselines plateau well below the human reference. 
We therefore plot human performance as a horizontal reference rather than as a full trajectory, with dotted vertical lines marking the mean observed human dialogue length; Appendix Table~\ref{tab:avg_dialogue_lengths} reports the realized average dialogue length for each AI condition. This framing also clarifies the early-turn region, where human clinicians often begin with greeting and rapport-building turns whereas AI agents move more directly into task-focused questioning.

\subsection{Why Equal Aggregate Success Masks Different Behavior}
\label{sec:outcome_conditioned_behavior}
Doctor-R1 at 20 turns and the human reference reach the same aggregate intervention Success (0.427), but they do not succeed on the same cases. Pairing all 300 cases yields 63 both-succeed cases, 65 human-only successes, 65 Doctor-R1-only successes, and 107 both-fail cases. We therefore stratify the paired dialogues by outcome and use the shared turn evaluator to summarize clinician intent and empathy.

\begin{table}[t]
\centering
\small
\setlength{\tabcolsep}{4pt}
\renewcommand{\arraystretch}{1.06}
\begin{tabular}{@{}lrrrr@{}}
\toprule
Outcome cell & $n$ & Concern probe (\%) & Reassure+plan (\%) & Turns \\
& & H / D & H / D & H / D \\
\midrule
Both succeed & 63 & 16.8 / 10.7 & 32.1 / 61.4 & 7.2 / 11.2 \\
Human only & 65 & 19.2 / 0.8 & 33.2 / 89.4 & 7.9 / 19.3 \\
Doctor-R1 only & 65 & 6.0 / 12.6 & 42.9 / 60.2 & 8.8 / 11.2 \\
Both fail & 107 & 5.7 / 1.3 & 55.3 / 89.8 & 8.7 / 18.5 \\
\bottomrule
\end{tabular}
\caption{Outcome-conditioned intervention behavior for humans (H) and Doctor-R1 at 20 turns (D). Intent percentages are shares of clinician turns labeled as concern probes or as reassurance/plan instruction.}
\label{tab:outcome_conditioned_behavior}
\end{table}

The human-only cell exposes a \emph{prescribe-before-probing} failure: Doctor-R1 spends 89.4\% of turns on reassurance or plan instruction but only 0.8\% on concern probes, despite using 19.3 turns versus 7.9 for humans. The direction reverses in Doctor-R1-only cases, where its extra turns include more concern probing than the human dialogue (12.6\% versus 6.0\%). Across all intervention turns, Doctor-R1 is also labeled moderate-or-strong in empathy more often than humans (75.9\% versus 23.4\%), showing that judged warmth can coexist with a generic, concern-agnostic plan. In a representative human-only case involving a knee injury and a remote-island access barrier, the clinician elicits cost and travel constraints and proposes telehealth; Doctor-R1 recommends orthopedic evaluation from the outset but never surfaces the access concern. The actionable lesson is therefore not simply to extend the conversation: clinician agents should probe before prescribing and condition the plan on the concern actually elicited.

\subsection{Additional Robustness and Coverage Analyses}

Evaluator checks yielded substantial human agreement ($\kappa=0.75$) and preserved confirmation ordering across all tested policy configurations; intervention ordering held in four of five, with the sole reversal at a near-zero floor, and an evaluator-backbone swap preserved both task orderings (Appendix Sections~\ref{sec:appendix_evaluator_validation}--\ref{sec:appendix_state_sensitivity}). Concern-type breakdowns show that the human reference exceeds the strongest matched-length model on three of four types in both tasks, with similar human--model patterns in the interpretable severity strata (Appendix Section~\ref{sec:appendix_concern_breakdowns}). Human outcomes are stable by experience; first-language groups differ only in confirmation Reveal Rate, and aggregate outcomes largely plateau by turn 12 (Appendix Section~\ref{sec:appendix_human_robustness}).

%% file: sections/06_conclusion.tex
\section{Conclusion}

We presented \benchmark, a benchmark and reserved patient simulator for hidden-concern reasoning in medical dialogue under partial observability. Built from real-world clinician-responded online health discussions, \benchmark separates visible clinical information from simulator-held hidden concerns and evaluates two abilities: \emph{confirmation} (eliciting concerns through dialogue) and \emph{intervention} (addressing them to guide patient decision).
Across experiments, no single AI system performs best across both tasks. Human clinicians remain strongest in elicitation and efficient intervention, while AI systems exhibit complementary strengths and benefit from longer dialogues at the cost of efficiency. 
These results highlight that effective medical dialogue requires grounded interactive reasoning (eliciting, timing, and acting on patient concerns) rather than relying on static knowledge alone. More broadly, \benchmark provides a controlled testbed for NLP research on asymmetric-information interaction, while offering the healthcare domain a tool to evaluate and improve clinician-facing AI systems in realistic, communication-critical settings.

The benchmark remains a controlled evaluation rather than external clinical validation. Hidden concerns are reconstructed from patient-authored online text rather than prospectively patient-confirmed, the intervention task uses a single source-derived target plan, and outcomes depend on simulator state transitions. We validate those transitions through a blind sampled audit and sensitivity analyses, but prospective studies with standardized patients or patient-confirmed concerns remain necessary (Appendix Section~\ref{sec:appendix_scope_limitations}).

%% file: sections/05_related_work.tex
\section{Related Work}
\label{sec:appendix_related_work}

\paragraph{Medical Dialogue Benchmarks and Patient Simulation.}
Recent medical dialogue benchmarks and simulators have made LLM-based clinician evaluation substantially more interactive than static QA. AI Hospital develops a multi-agent medical interaction simulator for evaluating clinical decision-making in dialogue \citep{fan2025ai}, while PatientSim emphasizes persona-driven patient simulation for realistic clinician--patient exchanges \citep{kyungpatientsim}. More recent work also proposes rubric-based multi-turn consultation evaluation such as MedDialogRubrics \citep{gong2026meddialogrubrics}, and AI-generated patient simulators for counseling-oriented settings such as motivational interviewing and mental health training \citep{yosef2024motivational,wang2024patient}. These settings improve realism, but they generally do not center the specific asymmetric-information problem that motivates our benchmark: clinically important psychosocial concerns that are intentionally private and should become available only after appropriate elicitation. Our benchmark differs by explicitly reserving hidden concerns inside the patient state, separating chart-visible context from simulator-internal barriers, and evaluating both \emph{confirmation} (eliciting those concerns) and \emph{intervention} (addressing the most salient one and guiding plan acceptance).

\paragraph{LLM Clinical Inquiry and Clinician Agents.}
Several recent works focus on improving LLM question-asking and inquiry behavior in medical settings. MedIQ benchmarks interactive clinical reasoning through question asking \citep{li2024mediq}, and ALFA studies how to align language models to ask better clinical questions \citep{lialfa}. Doctor-R1 further pushes this direction by training clinical inquiry behavior with experiential reinforcement learning \citep{lai2025doctor}, while clinical assistant systems such as CLARITY target downstream routing, consultation, and triage decisions \citep{shaposhnikov2025clarity}. These efforts share our interest in interaction quality, but they target somewhat different failure modes. Our benchmark is less about general diagnostic questioning and more about hidden-concern communication: whether a clinician can uncover patient-held fears, misconceptions, communication barriers, or practical constraints, and then address the most salient barrier in a way that changes the downstream decision trajectory.

\paragraph{Clinical Communication Frameworks and Intervention Quality.}
Our turn-level signal design is grounded in established clinical communication frameworks rather than a new manual coding scheme. RIAS provides a broad coding vocabulary for clinician--patient interaction function and socio-emotional behavior \citep{roter2002roter}; VR-CoDES focuses on concern cues, emotional disclosure, and clinician response space \citep{del2017verona}; and the Necessity--Concerns Framework links patient beliefs and worries to treatment adherence \citep{horne2013understanding}. Prior work on adherence-promotion skills, medication nonadherence communication, and patient-centered communication in treatment decision making further motivates barrier-targeting, explanation quality, and actionability as clinically meaningful interaction dimensions \citep{schlundt1994evaluation,wilson2007physician,cuevas2019mistrust}. Motivational interviewing evaluation is also relevant here because it emphasizes eliciting latent barriers and supporting patient change talk through dialogue rather than one-shot recommendation delivery \citep{yosef2024motivational}. Recent work on value-framed health communication similarly motivates our value-detection analysis \citep{heine2021using,winters2024moral}. We build on this literature by operationalizing these ideas as shared turn-level signals for both hidden-state transitions and post-hoc communication analysis.

\paragraph{Interactive Evaluation under Partial Observability.}
Outside medicine, recent work has emphasized evaluating LLM agents in settings with asymmetric or private information. Benchmarks for communication and verification in collaborative agents \citep{peng2025communication}, bargaining and negotiation under hidden preferences \citep{xia2024measuring,yang2021personality}, interaction-based alignment evaluation such as CUPID \citep{kimcupid}, and planning-theory-of-mind persuasion tasks such as MINDGAMES \citep{moorelarge} all show that static single-turn evaluation misses important aspects of interactive competence. These works also motivate more human-grounded and interaction-grounded evaluation of persuasion and alignment rather than relying only on single-model self-report or final-answer accuracy. Our benchmark brings this perspective into patient--clinician communication, where the hidden state is not a game payoff but a clinically meaningful psychosocial barrier. This motivates our emphasis on reserved patient state, process-grounded concern matching, and matched human--AI comparison under the same observable information. Compared with purely judge-based evaluation designs, our main task outcomes are anchored in simulator state transitions and trace-derived process signals, while the LLM-judge style analysis remains a complementary layer that could be paired with direct human rating in future robustness studies.

%% file: sections/07_appendix.tex
\section{Data and Case Construction}

\subsection{Source Data and Conversation Extraction}
We downloaded submissions and comments from \texttt{r/AskDocs} using the Arctic Shift download tool, covering July 11, 2013 to January 20, 2026. We retained threads marked with the subreddit flair \texttt{Physician Responded} and extracted alternating physician--patient chains with at least 8 turns (i.e., 4 back-and-forth exchanges), beginning from a clinician-flair comment and following a single reply chain for downstream case construction. We imposed this minimum-length requirement because longer exchanges are more likely to surface follow-up questions, clarification, and psychosocial context that are useful for constructing reserved patient profiles. On \texttt{r/AskDocs}, flaired medical professionals are moderator-verified, so this filter increases the likelihood that the retained conversations reflect interactions with medically credentialed responders.

\subsection{Common-Condition Subset Construction}
From the extracted pool, we constructed a common-condition subset so that clinicians could focus on confirmation and intervention quality rather than being disproportionately penalized by very rare or highly specialized presentations. We first identified eligible candidates and then ranked them with a GPT-based commonness scoring procedure, where lower rarity scores correspond to more common presentations. We retained the top 300 cases for downstream profile generation and benchmark subset construction.

\subsection{Patient Profile Taxonomy} \label{patient_taxonomy}
Each benchmark case is represented as a canonical JSON patient profile that serves as the source of truth for downstream simulation. The profile includes identifiers a \texttt{demographics} block (name, age, sex, marital status, education, background); a \texttt{clinical} block (admission reason, adherence behavior, medical/surgical history); and a \texttt{psychosocial} block (personality, current life situation, family medical history, lifestyle history, family dynamics). To support the hidden-information design, each profile also contains a list of \texttt{hidden\_concerns}, where every concern is represented by free-text content, a canonical category label, a confidence score, and short evidence snippets grounded in the source conversation. Beyond patient description, the profile stores an \texttt{intervention} block identifying the most clinically relevant hidden concern and summarizing the patient's original preference and recommended plan, a \texttt{roleplay} block specifying response style and disclosure behavior, and \texttt{self\_management\_domains} describing awareness, adherence, communication, and regimen execution.

\subsection{Patient Case Population}
\label{sec:appendix_patient_population}
The final benchmark subset contains 300 patient cases. To characterize the clinical breadth of this subset, we assigned each case a single best-fit specialty label under the clinician-role taxonomy used elsewhere in the study. All 300 cases received a valid specialty assignment. Because that taxonomy does not enumerate every clinically relevant subspecialty, we report the previously overloaded categories ``Other Specialist Doctor'' and ``Other Specialist Surgeon'' as \emph{unmapped medical specialty} and \emph{unmapped surgical specialty}, respectively. These buckets capture cases that were better described by specialties not explicitly listed in the study taxonomy, such as infectious disease, pulmonology, rheumatology, otolaryngology, and obstetrics/gynecology.

Table~\ref{tab:case_specialty_distribution} shows the resulting specialty distribution. The benchmark is not concentrated in a single field: the most frequent groups are unmapped medical specialty ($n=51$), gastroenterology ($n=40$), family medicine / general practice ($n=29$), dermatology ($n=28$), cardiology ($n=22$), psychiatry ($n=20$), and unmapped surgical specialty ($n=17$). The remaining cases are distributed across emergency medicine, ophthalmology, urology, orthopedics, surgery, neurology, hematology, pediatrics, nephrology, plastic surgery, neurosurgery, oncology, radiology, geriatrics, and vascular surgery.

{\scriptsize
\setlength{\tabcolsep}{4pt}
\renewcommand{\arraystretch}{1.08}
\begin{table}[!t]
\centering
\begin{tabular}{p{0.72\linewidth}r}
\toprule
\textbf{Best-fit specialty} & \textbf{Cases} \\
\midrule
Unmapped medical specialty & 51 \\
Gastroenterologist & 40 \\
Family Physician / General Practitioner & 29 \\
Dermatologist & 28 \\
Cardiologist & 22 \\
Psychiatrist & 20 \\
Unmapped surgical specialty & 17 \\
Emergency Medicine Specialist & 15 \\
Ophthalmologist & 12 \\
Urologist & 12 \\
Orthopedic Surgeon & 11 \\
General Surgeon & 8 \\
Neurologist & 7 \\
Hematologist & 6 \\
Pediatrician (Paediatrician) & 6 \\
Nephrologist & 4 \\
Plastic / Reconstructive Surgeon & 4 \\
Neurosurgeon & 2 \\
Oncologist & 2 \\
Radiologist & 2 \\
Geriatrician & 1 \\
Vascular Surgeon & 1 \\
\bottomrule
\end{tabular}
\caption{Best-fit specialty distribution for the 300-case benchmark subset. ``Unmapped'' categories denote clinically appropriate specialties not explicitly represented in the study's clinician-role taxonomy.}
\label{tab:case_specialty_distribution}
\end{table}
}

\subsection{Annotation and Labeling}
Patient profile construction required structured annotation over the source threads. During this process, clinically salient facts, psychosocial context, and hidden concerns were extracted from the clinician--patient exchange and organized into the canonical profile schema described above. Hidden concerns were then normalized into the fixed category set used throughout the benchmark so that profile generation, patient simulation, and evaluation all operate over the same label space.

To support Task 2 intervention construction, we additionally applied a broad multi-label scenario classification step over the retained clinician-responded conversations. Using an internal clinical communication taxonomy and guidance from a clinician researcher on the study team, we prioritized only conversations falling into two intervention-oriented families: \emph{Goals of Care, Prognostication, and End-of-Life Communication} and \emph{Misinformation, Distrust, and Alternative Belief Systems}. We used this filtering step to identify cases for which a clinically meaningful intervention target could be specified, including the patient's initial preference, the recommended plan, and the key hidden barrier whose resolution should move the interaction toward that plan. The broader internal framework was used only for internal case triage and intervention-goal design, and is therefore not reproduced in full here.

\section{Human Clinician Study Details}
\label{sec:appendix_human_clinician}
\subsection{Recruitment and Eligibility Criteria}
We used Prolific's built-in screener filters to recruit clinician participants. Specifically, we required the screener field \emph{Current Job Role} to match one of the following:
\begin{itemize}
    \item Cardiologist; Cardiothoracic Surgeon; Dermatologist; Emergency Medicine Specialist; Family Physician / General Practitioner; Gastroenterologist; General Medical Doctor (non-specialist); General Surgeon; Genetic Counselor; Geriatrician; Hematologist; Nephrologist; Neurologist; Neurosurgeon; Oncologist; Ophthalmologist; Orthopedic Surgeon; Other Specialist Doctor; Other Specialist Surgeon; Pediatrician (Paediatrician); Plastic / Reconstructive Surgeon; Podiatrist / Chiropodist; Psychiatrist; Radiologist; Urologist; Vascular Surgeon.
\end{itemize}
We additionally required the Prolific screener field \emph{Fluent Languages} to be English. All remaining participant descriptors reported in this appendix (survey responses and demographics) are post-recruitment sample-characterization variables rather than additional eligibility rules.

\subsection{Study Procedure and Interface}
Participants progressed through the study in a fixed interface sequence. Before full deployment, we iterated on this procedure through pilot runs with a clinician collaborator and a small initial Prolific sample, using those pilots to refine wording, interface flow, and the final task timing. First, participants viewed the consent form in an embedded PDF viewer and recorded an explicit agree/disagree decision before entering the study; the materials informed them that the patient was AI-simulated. After consent, they completed the brief background survey and were then routed to a patient-chart review page. The chart page displayed the assigned task type, overall progress, and an EMR-style summary containing patient description, encounter reason, and medical/surgical history. From this page, participants launched either a confirmation interview or an intervention interview, depending on their assignment.

Both interview pages used a chat-based interface with text entry, optional voice input, a persistent patient-chart sidebar, and a visible 10-minute countdown timer with a reminder at 2 minutes remaining. In the confirmation page, participants conducted the dialogue and then submitted structured findings; the interface required at least five clinician turns before findings could be submitted and allowed multiple concern entries consisting of a category label and free-text description. In the intervention page, participants were shown the patient's initial preference and the target clinical goal, and the session remained active until either the patient accepted the proposed plan or the timer expired. Early manual termination of intervention was enabled only after successful acceptance.

\subsection{Study Platform Screenshots}
\label{sec:appendix_platform_screenshots}
\begin{figure*}[t]
\centering
\includegraphics[width=0.9\textwidth,height=0.78\textheight,keepaspectratio]{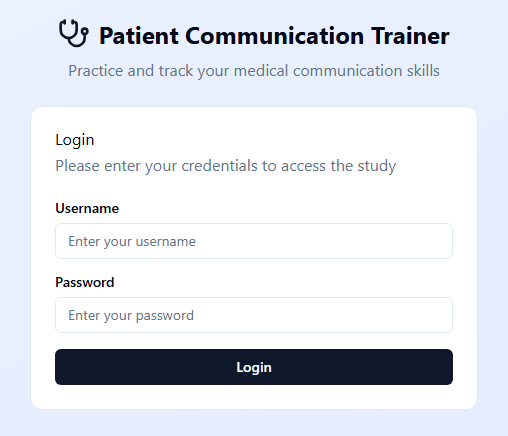}
\caption{Login screen for the clinician study platform.}
\label{fig:platform_login}
\end{figure*}

\begin{figure*}[t]
\centering
\includegraphics[width=0.9\textwidth,height=0.78\textheight,keepaspectratio]{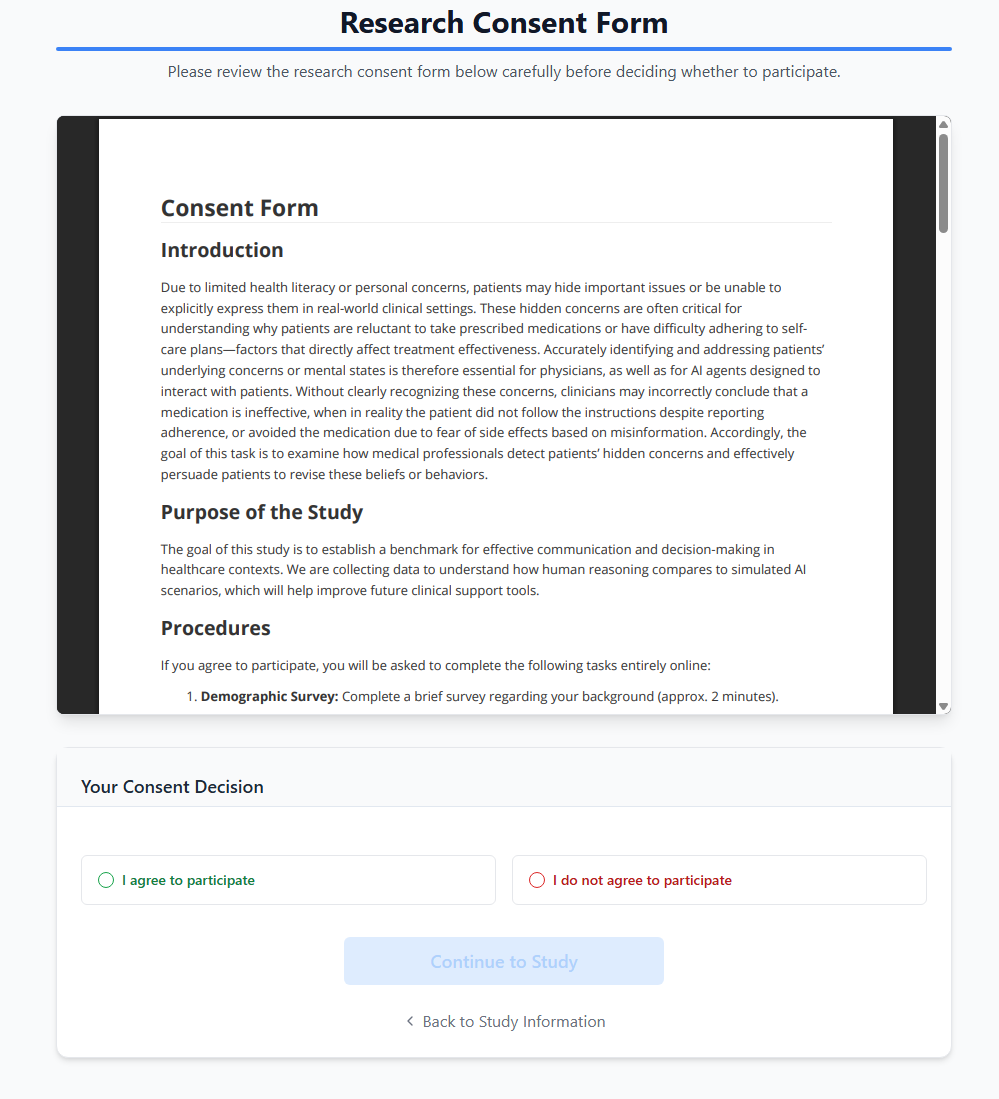}
\caption{Consent-form screen shown before study entry.}
\label{fig:platform_consent}
\end{figure*}

\begin{figure*}[t]
\centering
\includegraphics[width=0.9\textwidth,height=0.78\textheight,keepaspectratio]{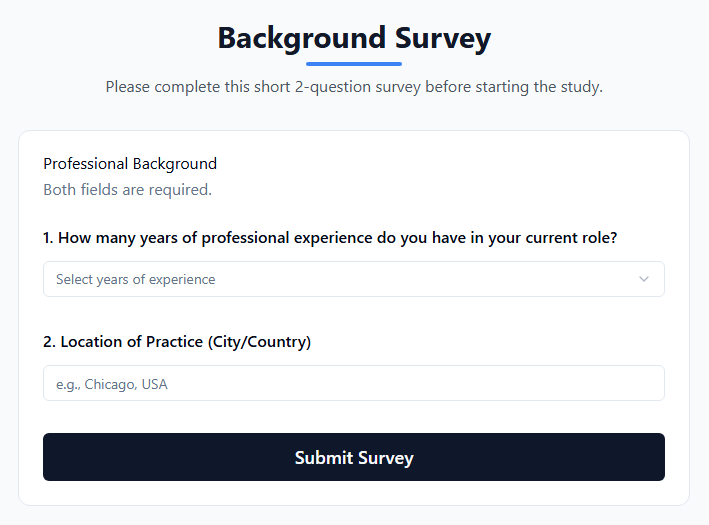}
\caption{Background-survey screen completed before the study tasks.}
\label{fig:platform_survey}
\end{figure*}

\begin{figure*}[t]
\centering
\includegraphics[width=0.9\textwidth,height=0.78\textheight,keepaspectratio]{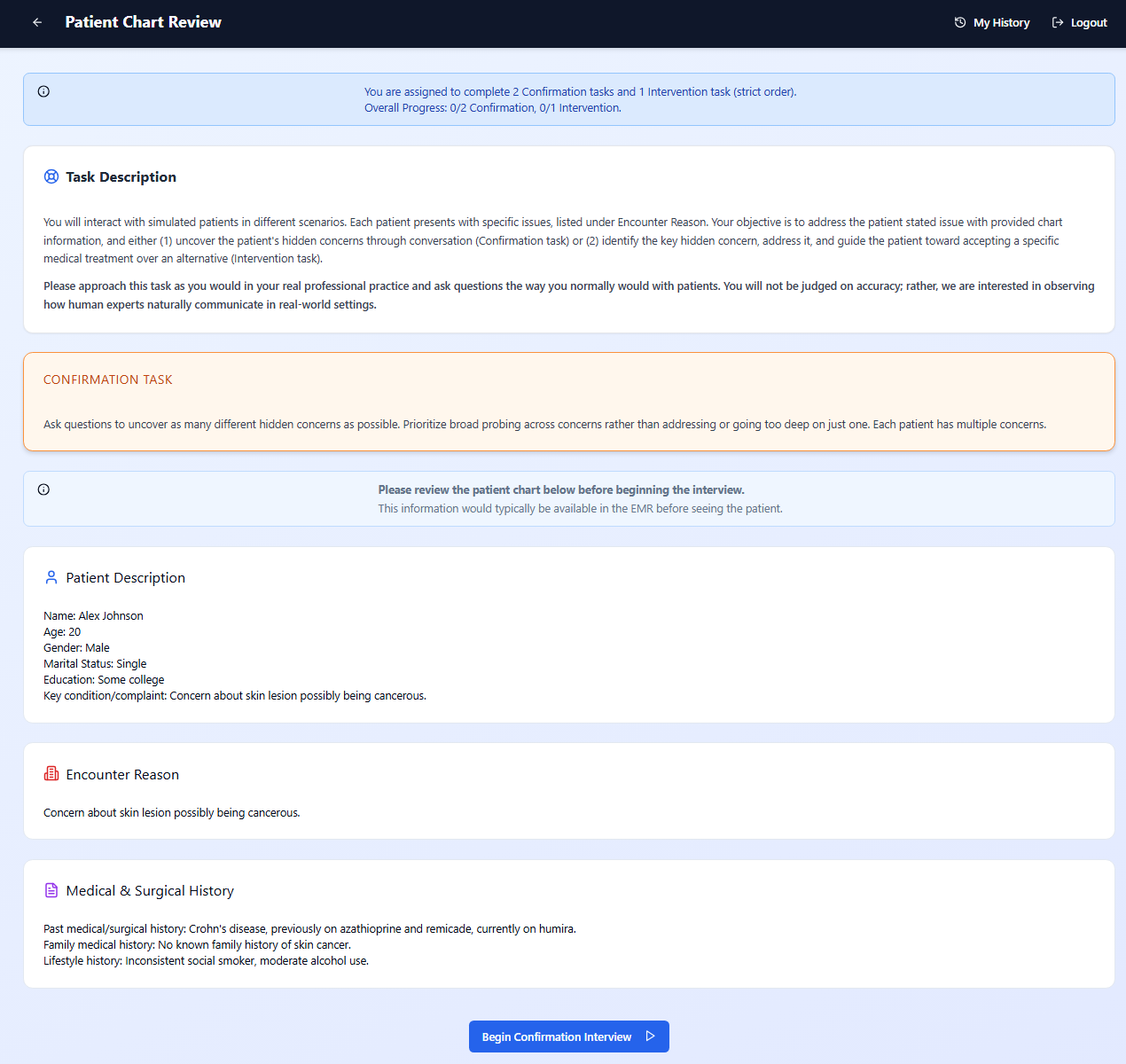}
\caption{Confirmation-task chart view, which presents the clinician-visible case summary before the dialogue starts.}
\label{fig:platform_confirmation_chart}
\end{figure*}

\begin{figure*}[t]
\centering
\includegraphics[width=0.9\textwidth,height=0.78\textheight,keepaspectratio]{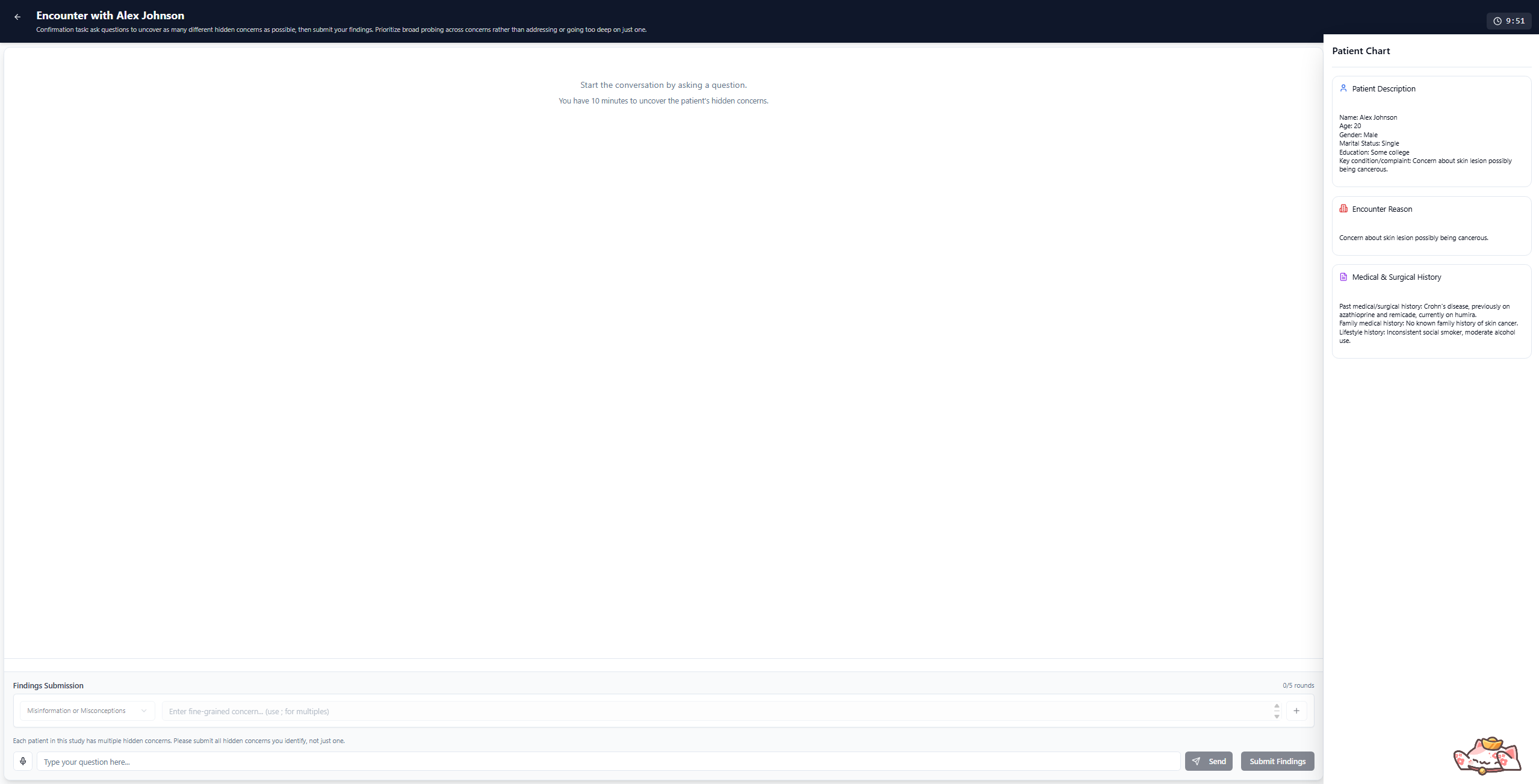}
\caption{Confirmation-task dialogue page used for the chat-based interview and structured concern submission.}
\label{fig:platform_confirmation_page}
\end{figure*}

\begin{figure*}[t]
\centering
\includegraphics[width=0.9\textwidth,height=0.78\textheight,keepaspectratio]{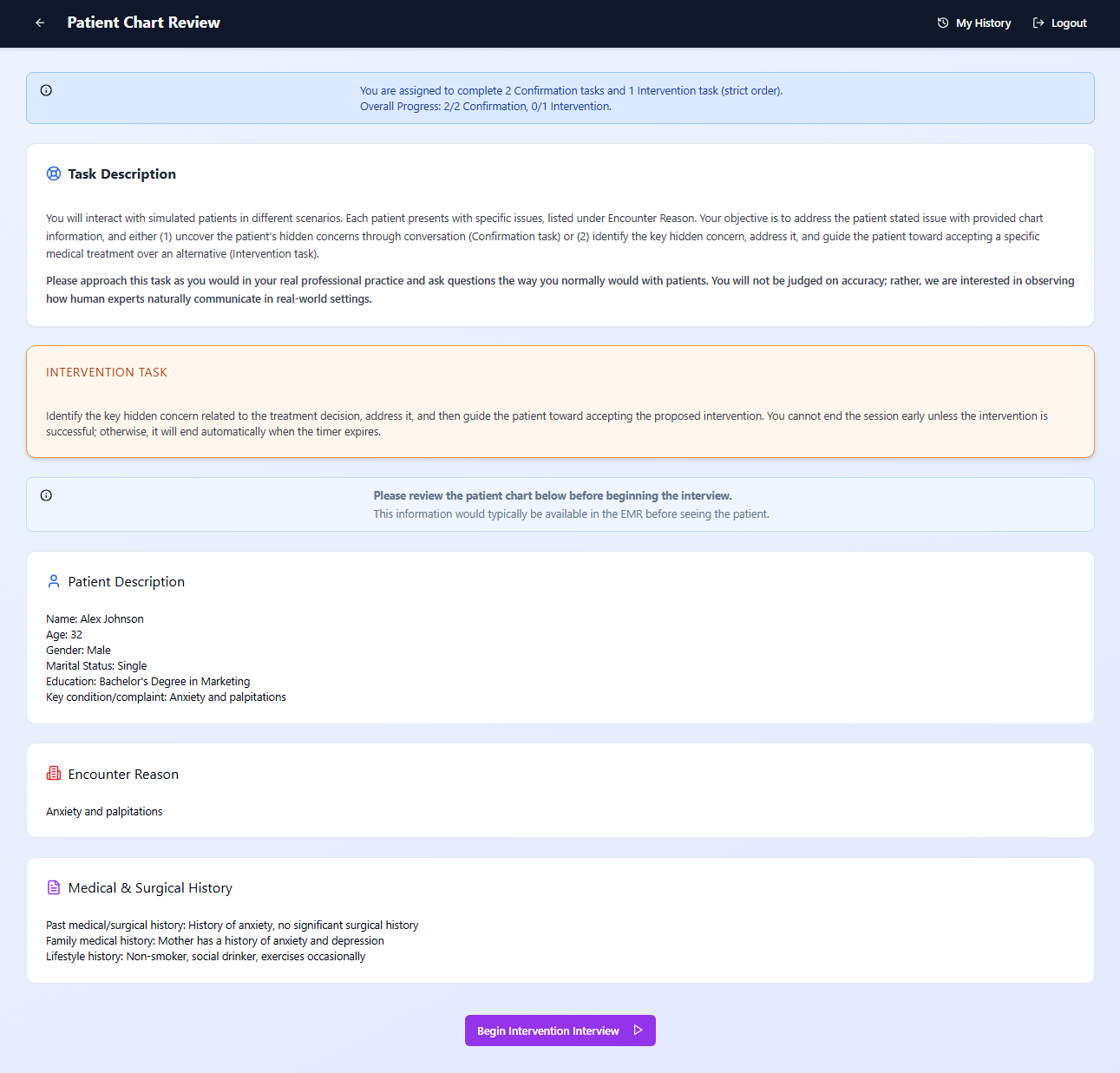}
\caption{Intervention-task chart view, which presents the clinician-visible case summary together with the initial preference and target plan.}
\label{fig:platform_intervention_chart}
\end{figure*}

\begin{figure*}[t]
\centering
\includegraphics[width=0.9\textwidth,height=0.78\textheight,keepaspectratio]{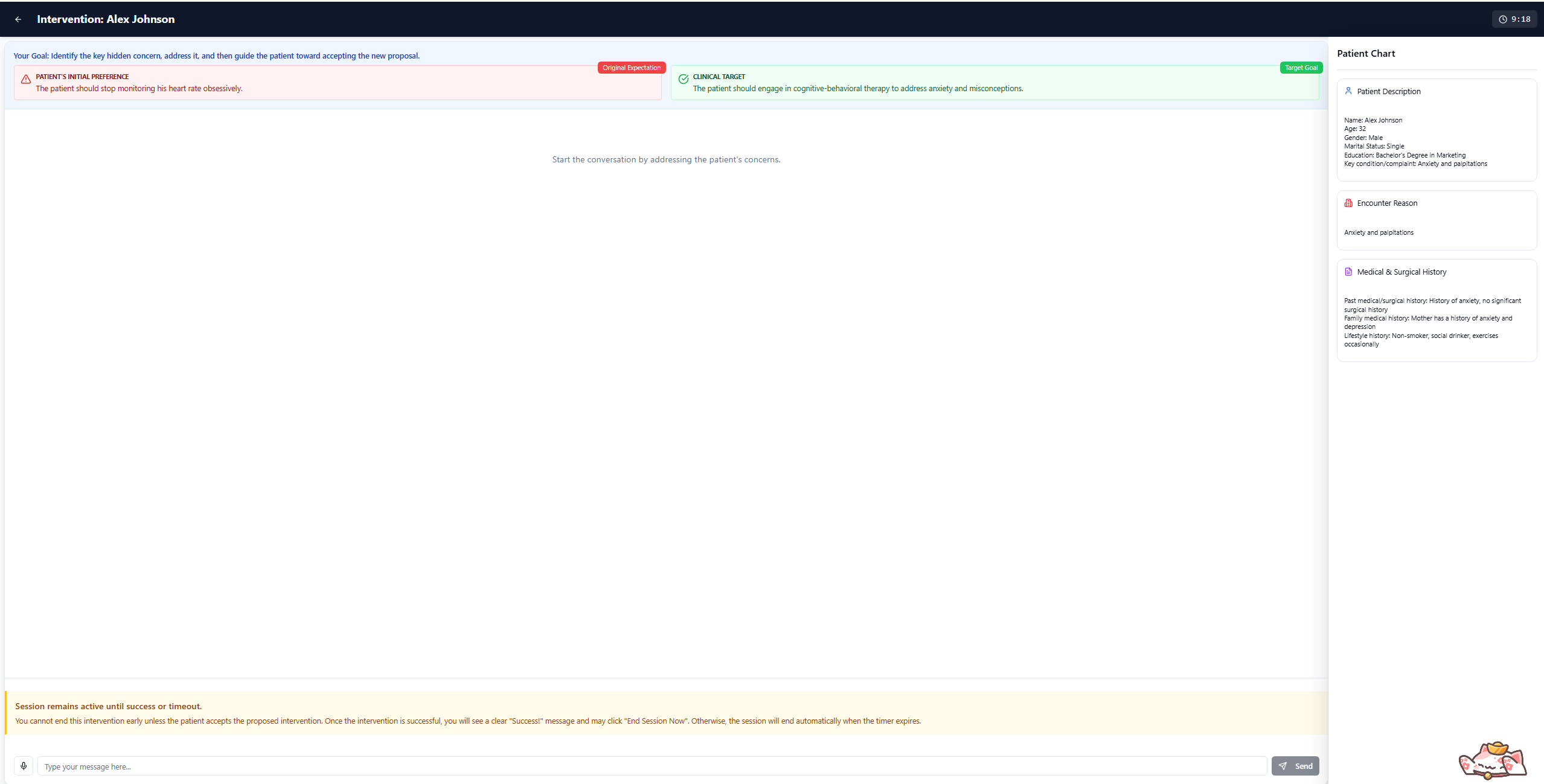}
\caption{Intervention-task dialogue page used until acceptance or timeout.}
\label{fig:platform_intervention_page}
\end{figure*}

Figures~\ref{fig:platform_login}--\ref{fig:platform_intervention_page} show the human-study platform in the order participants encountered it: login, consent, survey, confirmation chart review, confirmation dialogue, intervention chart review, and intervention dialogue.

\subsection{Instructions and Materials}
The participant-facing instructions were integrated into the interface rather than provided as a separate manual. On the chart review page, participants were told to address the patient's stated issue using the available chart information and either (1) uncover hidden concerns through conversation (confirmation) or (2) identify the key hidden concern, address it, and guide the patient toward accepting a specific medical treatment over an alternative (intervention). They were also instructed to approach the interaction as they would in real professional practice, and informed that the study was intended to observe natural expert communication rather than judge factual accuracy.

Task-specific prompts appeared again at encounter launch. For confirmation, participants were instructed to ask questions that uncovered as many different hidden concerns as possible, to prioritize broad probing rather than going too deep on a single issue, and to submit all hidden concerns they identified. The confirmation interface also stated that each patient had multiple hidden concerns. For intervention, participants were instructed to identify the key hidden concern, address it, and guide the patient toward the proposed intervention; the interface explicitly stated that the session could not be ended early unless the patient accepted the plan. Supporting materials included the chart summary for both tasks, the structured concern-entry form for confirmation, and the paired display of the patient's initial preference and target clinical goal for intervention.

\subsection{Survey Instrument}
Before starting the study tasks, participants completed a brief background survey with two required items. The first asked for years of professional experience in the participant's current role, using ordinal response options ranging from \emph{Student / trainee} to \emph{More than 20 years}. The second asked for location of practice in free-text form (city/country). We used these responses to summarize clinician background and to contextualize the clinician-participant sample.

\subsection{Demographics}
In addition to the brief background survey, we retained available participant demographic variables exported from Prolific for sample description. Percentages below are reported over the available responses for each field and are used only to describe the clinician-participant sample in aggregate.

\paragraph{Age.}
The clinician sample had mean age 31.4, median age 31, and range 20--71.

\paragraph{Sex.}
Available sex metadata indicated 77 male clinicians (68.1\%), 35 female clinicians (31.0\%), and 1 prefer-not-to-say response (0.9\%).

\paragraph{Clinical role.}
Reported clinical roles were general medical doctor (non-specialist), 41.6\%; family physician / general practitioner, 15.0\%; other specialist doctor, 13.3\%; emergency medicine specialist, 10.6\%; radiologist, 2.7\%; oncologist, 2.7\%; orthopedic surgeon, 1.8\%; general surgeon, 1.8\%; psychiatrist, 1.8\%; dermatologist, 1.8\%; other specialist surgeon, 1.8\%; pediatrician, 0.9\%; neurologist, 0.9\%; hematologist, 0.9\%; urologist, 0.9\%; cardiothoracic surgeon, 0.9\%; and cardiologist, 0.9\%.

\paragraph{Ethnicity.}
Reported ethnicity was White, 47.8\%; Asian, 15.9\%; Black, 11.5\%; Mixed, 11.5\%; Other, 11.5\%; and prefer-not-to-say, 1.8\%.

\paragraph{Country of residence.}
Reported countries of residence were Egypt, 23.9\%; United Kingdom, 19.5\%; United States, 12.4\%; Canada, 5.3\%; Mexico, 5.3\%; South Africa, 3.5\%; Brazil, 3.5\%; Italy, 3.5\%; Spain, 3.5\%; India, 3.5\%; Argentina, 2.7\%; Netherlands, 1.8\%; Portugal, 1.8\%; Slovenia, 1.8\%; Australia, 0.9\%; Chile, 0.9\%; Germany, 0.9\%; Greece, 0.9\%; Ireland, 0.9\%; Morocco, 0.9\%; Poland, 0.9\%; Saudi Arabia, 0.9\%; and Sweden, 0.9\%.

\paragraph{Years of experience.}
Reported years of experience were student / trainee, 2.0\%; less than 1 year, 4.9\%; 1--2 years, 28.8\%; 3--5 years, 36.1\%; 6--10 years, 15.1\%; and 11--20 years, 10.2\%.

\subsection{Average Dialogue Lengths}
Table~\ref{tab:avg_dialogue_lengths} reports the average conversation lengths used to contextualize the main-paper turn-accumulation plots. Human rows are the observed clinician conversations collected in the study, whereas AI rows are the realized mean lengths of the corresponding fixed or adaptive runs. These averages are descriptive only: they summarize how long each condition typically runs, not a target that every conversation is forced to match.

{\small
\setlength{\tabcolsep}{6pt}
\renewcommand{\arraystretch}{1.08}
\begin{table}[htbp]
\centering
\begin{tabular}{lcc}
\toprule
Model / Condition & Confirmation & Intervention \\
\midrule
Human clinicians & 8.2 & 8.2 \\
Qwen-3.5-9B & 19.4 & 18.9 \\
Claude Sonnet 4.5 & 11.7 & 15.4 \\
GPT-5.2 & 11.7 & 18.6 \\
Doctor-R1 & 18.1 & 15.5 \\
Llama3-OpenBioLLM-8B & 15.5 & 19.5 \\
\bottomrule
\end{tabular}
\caption{Average realized dialogue length by model and task. Human values are the observed study averages; AI values are the realized averages of the corresponding benchmark runs.}
\label{tab:avg_dialogue_lengths}
\end{table}
}

\section{Evaluation Details}
\subsection{Concern Matching for Submitted Findings}
\label{sec:appendix_concern_matching}
Confirmation findings are evaluated with a shared post-hoc matching procedure applied identically to human submissions and AI post-dialogue extraction outputs. Each predicted finding contains a canonical category label and a short free-text concern description; gold concerns are read from the structured patient profile. Fine-grained matching is one-to-one: each predicted finding can match at most one gold concern, and each gold concern can be matched at most once. A fine-grained match is accepted only when the judge determines that the submitted concern is semantically correct at the specific concern level and is supported by explicit patient-stated evidence in the conversation. To enforce process grounding, the matcher is instructed not to credit a match based on chart context, likely diagnosis, or plausible clinician inference alone.

For every accepted fine-grained match, the judge also records whether the submitted category matches the gold category. This yields the fine-grained precision/recall/F1 metrics, conditional category accuracy over matched pairs, and category-aware fine-grained summaries used in the appendix diagnostics. Separately, we compute coarse-grained category metrics by collapsing both predicted findings and gold concerns to the four-category taxonomy and comparing category counts. Coarse-grained precision, recall, and F1 therefore measure whether the clinician recovered the correct \emph{types} of hidden barriers even when the exact fine-grained phrasing differs.

We additionally report process-grounded extraction metrics by restricting the eligible gold set to concerns that were actually revealed during the interaction under the patient simulator's state trace. Process-grounded precision credits submitted findings only when they match concerns that were surfaced in dialogue, and process-grounded recall is normalized by the number of revealed gold concerns rather than the total hidden-concern set. The matched-but-no-reveal diagnostic flags cases where at least one submitted finding matches a gold concern but no gold concern was explicitly revealed in the trace, thereby separating post-hoc guessing from successful elicitation.

\subsection{Pairwise Statistical Comparisons for Confirmation}
We additionally summarize case-level statistical comparisons for the confirmation metrics in Table~\ref{tab:level1_human_ai}. These appendix analyses operate on per-case metric values averaged over the 300 benchmark cases, whereas the main-paper table reports concatenated (micro-aggregated) counts before computing precision, recall, and F1. The reference values in this appendix therefore need not numerically match Table~\ref{tab:level1_human_ai}, and this difference is expected rather than an inconsistency. To keep the appendix focused, we report only metrics for which either human reference row in Table~\ref{tab:level1_human_ai} is better than the strongest AI comparator under the corresponding comparison family. Each row reports the mean difference $\Delta=\text{Ref}-\text{AI}$, its 95\% confidence interval, and the Holm-adjusted $p$-value within that family of confirmation metrics.

{\small
\setlength{\tabcolsep}{4pt}
\renewcommand{\arraystretch}{1.08}
\begin{table*}[htbp]
\centering
\begin{tabular}{llccccc}
\toprule
Metric & Best AI & Ref & AI & $\Delta$ & 95\% CI & Holm $p$ \\
\midrule
Fine R & Claude (Adaptive) & 0.649 & 0.555 & +0.094 & [+0.060, +0.129] & 0.0002*** \\
Fine F1 & Claude (Adaptive) & 0.570 & 0.513 & +0.058 & [+0.028, +0.087] & 0.0004*** \\
\bottomrule
\end{tabular}
\caption{Case-level comparison for the \emph{Claude summary} reference row in Table~\ref{tab:level1_human_ai} versus the best AI system under any confirmation setting. Only metrics with a human-reference advantage are shown.}
\label{tab:confirmation_stats_human_claude}
\end{table*}
}

{\small
\setlength{\tabcolsep}{4pt}
\renewcommand{\arraystretch}{1.08}
\begin{table*}[htbp]
\centering
\begin{tabular}{llccccc}
\toprule
Metric & Best AI & Ref & AI & $\Delta$ & 95\% CI & Holm $p$ \\
\midrule
Coarse P & GPT-5.2 (8-turn) & 0.854 & 0.705 & +0.149 & [+0.108, +0.190] & 0.0002*** \\
Fine P & GPT-5.2 (8-turn) & 0.788 & 0.623 & +0.165 & [+0.116, +0.212] & 0.0002*** \\
\bottomrule
\end{tabular}
\caption{Case-level comparison for the \emph{Human summary} reference row in Table~\ref{tab:level1_human_ai} versus the best AI system in the matched 8-turn confirmation comparison. Only metrics with a human advantage are shown.}
\label{tab:confirmation_stats_human}
\end{table*}
}

{\small
\setlength{\tabcolsep}{4pt}
\renewcommand{\arraystretch}{1.08}
\begin{table*}[htbp]
\centering
\begin{tabular}{llccccc}
\toprule
Metric & Best AI & Ref & AI & $\Delta$ & 95\% CI & Holm $p$ \\
\midrule
Fine P & GPT-5.2 (Adaptive) & 0.788 & 0.663 & +0.125 & [+0.080, +0.169] & 0.0002*** \\
\bottomrule
\end{tabular}
\caption{Case-level comparison for the \emph{Human summary} reference row in Table~\ref{tab:level1_human_ai} versus the best AI system under any confirmation setting. Only metrics with a human advantage are shown.}
\label{tab:confirmation_stats_human_any}
\end{table*}
}

{\small
\setlength{\tabcolsep}{4pt}
\renewcommand{\arraystretch}{1.08}
\begin{table*}[htbp]
\centering
\begin{tabular}{llccccc}
\toprule
Metric & Best AI & Ref & AI & $\Delta$ & 95\% CI & Holm $p$ \\
\midrule
Coarse R & Claude (8-turn) & 0.746 & 0.630 & +0.115 & [+0.087, +0.142] & 0.0002*** \\
Coarse F1 & Claude (8-turn) & 0.657 & 0.620 & +0.037 & [+0.013, +0.060] & 0.0042** \\
Fine R & Claude (8-turn) & 0.649 & 0.515 & +0.134 & [+0.101, +0.167] & 0.0002*** \\
Fine F1 & Claude (8-turn) & 0.570 & 0.509 & +0.061 & [+0.032, +0.090] & 0.0002*** \\
\bottomrule
\end{tabular}
\caption{Case-level comparison for the \emph{Claude summary} reference row in Table~\ref{tab:level1_human_ai} versus the best AI system in the matched 8-turn confirmation comparison. Only metrics with a human-reference advantage are shown.}
\label{tab:confirmation_stats_human_claude_8turn}
\end{table*}
}

For the metrics reported in this subsection, the two human reference rows remain stronger than the best AI comparator on complementary parts of confirmation. The original human summaries retain the strongest precision-focused recovery, while the Claude-augmented summaries strengthen recall-oriented recovery and also improve matched-length coarse and fine F1 on the same elicitation traces.

\subsection{Pairwise Statistical Comparisons for Intervention}
We additionally summarize case-level statistical comparisons for the intervention metrics in Table~\ref{tab:level2_human_ai}. As in the confirmation appendix, these analyses operate on per-case metric values averaged over the 300 benchmark cases rather than on the concatenated main-paper aggregates. Because the main text treats the matched 8-turn condition as the fairness-aligned human--AI comparison, we focus on that family here and report only metrics for which the human reference is better than the strongest 8-turn AI comparator. Continuous metrics use paired permutation tests, Success uses an exact McNemar test, and confidence intervals are computed from paired bootstrap differences.

{\small
\setlength{\tabcolsep}{4pt}
\renewcommand{\arraystretch}{1.08}
\begin{table*}[htbp]
\centering
\begin{tabular}{llccccc}
\toprule
Metric & Best AI & Ref & AI & $\Delta$ & 95\% CI & Holm $p$ \\
\midrule
Reveal Rate & Claude (8-turn) & 0.487 & 0.313 & +0.173 & [+0.100, +0.247] & 0.0001*** \\
Success & Claude (8-turn) & 0.427 & 0.293 & +0.133 & [+0.060, +0.203] & 0.0004*** \\
\bottomrule
\end{tabular}
\caption{Case-level comparison for the \emph{Human clinicians} reference row in Table~\ref{tab:level2_human_ai} versus the best AI system in the matched 8-turn intervention comparison. Only metrics with a human advantage are shown.}
\label{tab:intervention_stats_human}
\end{table*}
}

The human intervention advantage is concentrated in the matched 8-turn comparison. Against the best AI system under any setting, human Reveal Rate remains numerically higher than Doctor-R1 at 20 turns (0.487 vs.\ 0.437), but the paired 95\% confidence interval for the difference overlaps zero ([-0.027, +0.127]), so no intervention metric shows a reliable human advantage once longer AI runs are allowed.

\subsection{Intervention Success Gate}
Intervention evaluation is defined by the patient simulator's internal state rather than by a separate final judge. For intervention cases, the benchmark retains a single primary target concern, derived from the profile's most clinically relevant hidden concern field and evaluated within the same fixed hidden-concern taxonomy used elsewhere in the benchmark.

At runtime, the target concern can occupy one of three statuses: hidden, revealed, or addressed. A case is counted as successful if the target concern reaches the addressed state at any turn in the dialogue. This gate is stricter than simply observing a persuasive or empathetic turn: the concern must first be revealed, the minimum reveal-to-address lag must be satisfied, and the addressing transition must fire under the same latent update rule used by the patient agent. The primary intervention outcome, Success, is therefore an observable projection of the simulator state, not a post-hoc annotator judgment.
We do not separately report a post-gate binary plan-acceptance variable in the current main-paper results. The evaluator retains the first addressed turn for timing analyses, but the success endpoint itself is simply whether the primary concern ever reaches the addressed state.

From the same trace, we derive intervention timing metrics. The first reveal turn is the earliest turn at which the target concern becomes revealed (or all remaining concerns are marked revealed), and the first addressed turn is the earliest turn at which the intervention status becomes addressed. We report both turn-to-addressed and reveal-to-address lag in the main paper, since these quantities remain comparable across fixed-length and success-capped intervention policies. Post-address diagnostics are retained only as appendix-level analyses: under the 20-turn success-capped intervention policy, runs stop immediately once success is reached, so post-address counts become structurally zero and post-address ratios may be undefined rather than behaviorally informative.

\subsection{Process Metrics from Rubric Signals}
Process metrics are computed directly from the per-turn trace emitted by the simulator and the shared evaluator. For every clinician turn, the runtime evaluator records intent, rubric features, and transition metadata in the \texttt{evaluator\_analysis} payload. This lets us derive stage-level summaries without re-running the simulator.

For confirmation, we compute open-versus-closed questioning with a separate judge that labels each clinician turn as open or closed based on whether it invites broad narrative explanation or seeks a short factual answer. Overall open-question ratio is the fraction of clinician turns labeled open, and early open-question ratio is the same quantity restricted to the first five turns. We also track elicitation dynamics from the reveal trace: full-reveal rate, turn-to-full-reveal, revealed-concern fraction by the turn limit, and average turns per revealed concern. These process measures are independent of the final submitted findings and therefore expose differences between conversational elicitation and post-dialogue extraction.

For safety and anti-gaming analysis, we use the evaluator's intent label and transition trigger. Meta-probe rate is the fraction of clinician turns classified as \emph{meta-category probing}, i.e., asking directly in benchmark label language rather than through natural clinical dialogue. Blocked-meta counts record how often such turns were prevented from changing the patient state by the anti-gaming rule. For intervention, we additionally compute appendix-only post-address diagnostics: the fraction of post-address turns that remain open-ended and the fraction of post-address patient turns that contain challenge cues such as continued worry, hesitation, or requests to avoid the plan. These metrics help distinguish clean closure from nominal success followed by renewed resistance, but they are only informative for traces that continue after the first addressed turn.

\subsection{Communication-Style Metrics}
\label{sec:appendix_style_metrics}
We report communication-style metrics as a complementary analysis layer rather than as part of the benchmark state-transition objective. These metrics are computed from transcript-level analyses and therefore should be interpreted as descriptive measures of \emph{how} the interaction is conducted, not whether the clinician succeeds under the hidden-state benchmark criteria. The style-analysis pipeline mixes deterministic text statistics with shared LLM-judge outputs, and is applied identically to human and AI clinician dialogues.

\paragraph{Information-structure features.}
The first family consists of deterministic transcript statistics computed from clinician turns only. In the main paper, we report Word/Turn, Readability, and Early Open. Readability is a standard formula, where larger values indicate easier-to-read language; by contrast, Word/Turn is reported as a descriptive verbosity measure rather than an objective to optimize. Readability metrics are standard formulas motivated by prior health-communication work on patient-facing materials \citep{daraz2018readability,wallace2004american}, while the turn-length and early-open statistics are project-defined descriptive measures.

\paragraph{Sender persuasive strategies.}
The second family scores sender-side strategy dimensions over the full clinician--patient conversation. In confirmation, the main paper reports Empathy, Collaboration, and Problem-Solving. In intervention, the main paper reports Empathy, Rationale, Problem-Solving, and Actionability. Each category is judged on a three-level scale ($0=\text{absent}$, $1=\text{present but partial}$, $2=\text{clear and effective}$). These dimensions are adapted from clinical counseling and adherence-promotion literature, especially prior work on training clinicians to diagnose barriers, explain treatment need, and support adherence behavior \citep{schlundt1994evaluation}, but the exact labels used here are benchmark-specific definitions rather than a direct reuse of an existing rubric.

\paragraph{Receiver value alignment.}
The third family evaluates whether the clinician recognizes values explicitly expressed by the patient. We use a project-defined clinical-values taxonomy with five categories: autonomy/control, safety/risk avoidance, trust/respect/dignity, family/social responsibility, and practical burden/cost/logistics. This taxonomy is inspired by prior work on value-framed health communication and moral messaging \citep{heine2021using,winters2024moral}, but is adapted to the clinician--patient setting rather than borrowed directly from moral-foundation taxonomies. In the main paper, we report Values Det., defined as the mean number of value categories explicitly expressed by patients and detected by the judge.

{\scriptsize
\setlength{\tabcolsep}{4pt}
\renewcommand{\arraystretch}{1.1}
\begin{table*}[htbp]
\centering
\begin{tabular}{p{0.18\linewidth} p{0.24\linewidth} p{0.22\linewidth} p{0.28\linewidth}}
\toprule
\textbf{Family} & \textbf{Main-paper metrics} & \textbf{Literature grounding} & \textbf{Interpretation} \\
\midrule
Information structure & Word/Turn, Readability, Early Open & Health-information readability \citep{daraz2018readability,wallace2004american} & Describes verbosity, readability, and early open-ended questioning without using hidden-state labels. \\
Sender persuasive strategies & Empathy, Collaboration (confirmation), Rationale (intervention), Problem-Solving, Actionability (intervention) & Clinical counseling and adherence-promotion skills \citep{schlundt1994evaluation} & Scores how well the clinician validates concerns, explains need, collaborates, addresses barriers, and gives actionable next steps. \\
Receiver value alignment & Values Det. & Value-framed health communication \citep{heine2021using,winters2024moral} & Measures whether the clinician recognizes patient values explicitly stated in the conversation. \\
  \bottomrule
  \end{tabular}
  \caption{Provenance of the communication-style metric families used in the main paper.}
\label{tab:style_metric_provenance}
\end{table*}
}

\subsection{Blind Validation of Evaluator-Derived State Transitions}
\label{sec:appendix_evaluator_validation}
Because the main outcomes are derived from the simulator state trace, we conducted an independent blind validation at both the per-turn event and complete-dialogue timing levels. For the event audit, we sampled 160 single-turn slices from human-reference dialogues: 80 reveal judgments and 80 address judgments, each balanced between positive and negative evaluator decisions. The independent annotator saw the dialogue context but not the evaluator label. This balanced sample is intentionally enriched for transition events and should not be interpreted as their natural prevalence.

{\small
\setlength{\tabcolsep}{6pt}
\renewcommand{\arraystretch}{1.08}
\begin{table}[htbp]
\centering
\begin{tabular}{lrrr}
\toprule
Judgment & $n$ & Cohen's $\kappa$ & Raw agreement \\
\midrule
Reveal & 80 & 0.725 & 0.863 \\
Address & 80 & 0.775 & 0.887 \\
Overall & 160 & 0.750 & 0.875 \\
\bottomrule
\end{tabular}
\caption{Blind agreement for per-turn reveal and address judgments.}
\label{tab:evaluator_turn_agreement}
\end{table}
}

For transition timing, the annotator reviewed 24 complete dialogues and marked the first reveal and, when applicable, first addressed turn. First-reveal timing matched exactly in 20/24 dialogues and within one turn in 22/24, with a mean absolute difference of 0.29 turns. Among the 12 intervention dialogues reviewed, both readings identified an addressed transition in 11; within those 11, the first-addressed turn matched exactly in 10, with a mean absolute difference of 0.18 turns. One dialogue had an addressed transition under only one reading. These results validate the quantities used by Reveal Rate, Success, and timing metrics on a sampled subset, but they rely on one independent annotator and do not establish prospective patient-level validity.

\subsection{State-Tracking and Disclosure Sensitivity}
\label{sec:appendix_state_sensitivity}
We tested whether the Doctor-R1 versus Qwen-3.5-9B comparison depends on a single latent-policy configuration. Because the simulator is closed-loop, each perturbation reruns all 300 AI dialogues end-to-end rather than merely rescoring frozen transcripts. The grid changes one policy component at a time: the reveal or address threshold, exponential-moving-average (EMA) smoothing, or concern-family deltas.

{\scriptsize
\setlength{\tabcolsep}{4pt}
\renewcommand{\arraystretch}{1.08}
\begin{table*}[htbp]
\centering
\resizebox{\textwidth}{!}{%
\begin{tabular}{llrrl}
\toprule
Task / metric & Configuration & Doctor-R1 & Qwen-3.5-9B & Ordering \\
\midrule
\multirow{4}{*}{Confirmation Coarse F1}
 & Anchor & 0.670 & 0.225 & Doctor-R1 $>$ Qwen \\
 & Reveal threshold $0.20\!\rightarrow\!0.25$ & 0.638 & 0.536 & Doctor-R1 $>$ Qwen \\
 & Reveal EMA $0.65\!\rightarrow\!0.55$ & 0.634 & 0.572 & Doctor-R1 $>$ Qwen \\
 & Concern-family deltas zeroed & 0.616 & 0.570 & Doctor-R1 $>$ Qwen \\
\midrule
\multirow{5}{*}{Intervention Success}
 & Anchor & 0.150 & 0.030 & Doctor-R1 $>$ Qwen \\
 & Reveal threshold $0.20\!\rightarrow\!0.25$ & 0.023 & 0.017 & Doctor-R1 $>$ Qwen \\
 & Address threshold $0.30\!\rightarrow\!0.25$ & 0.073 & 0.037 & Doctor-R1 $>$ Qwen \\
 & Address EMA $0.70\!\rightarrow\!0.60$ & 0.067 & 0.057 & Doctor-R1 $>$ Qwen \\
 & Concern-family deltas zeroed & 0.013 & 0.017 & Qwen $>$ Doctor-R1 \\
\bottomrule
\end{tabular}
}
\caption{End-to-end sensitivity to latent-policy choices. Confirmation ordering is preserved in every tested configuration. Intervention ordering is preserved in four of five configurations; zeroing concern-family deltas produces a small reversal at a near-zero floor.}
\label{tab:state_policy_sensitivity}
\end{table*}
}

The absolute intervention scores are more sensitive than the confirmation scores because Success requires both reveal and address transitions and can change subsequent patient behavior. Thus, the grid supports robustness of the main confirmation ordering but also confirms the AC's concern that intervention magnitudes depend on transition design. The one intervention reversal occurs when both systems score below 0.02 and does not support a claim of invariance across all policy settings.

We also replaced the per-turn evaluator backbone while holding the policy and post-hoc concern matcher fixed. With Claude Haiku 4.5 instead of \texttt{gpt-5-mini}, confirmation Coarse F1 is 0.619 for Doctor-R1 and 0.560 for Qwen, versus 0.670 and 0.225 at the anchor. Intervention Success is 0.103 and 0.077, versus 0.150 and 0.030 at the anchor. The ordering is preserved on both tasks, although the gap narrows.

\begin{figure*}[htbp]
\centering
\includegraphics[width=0.78\textwidth]{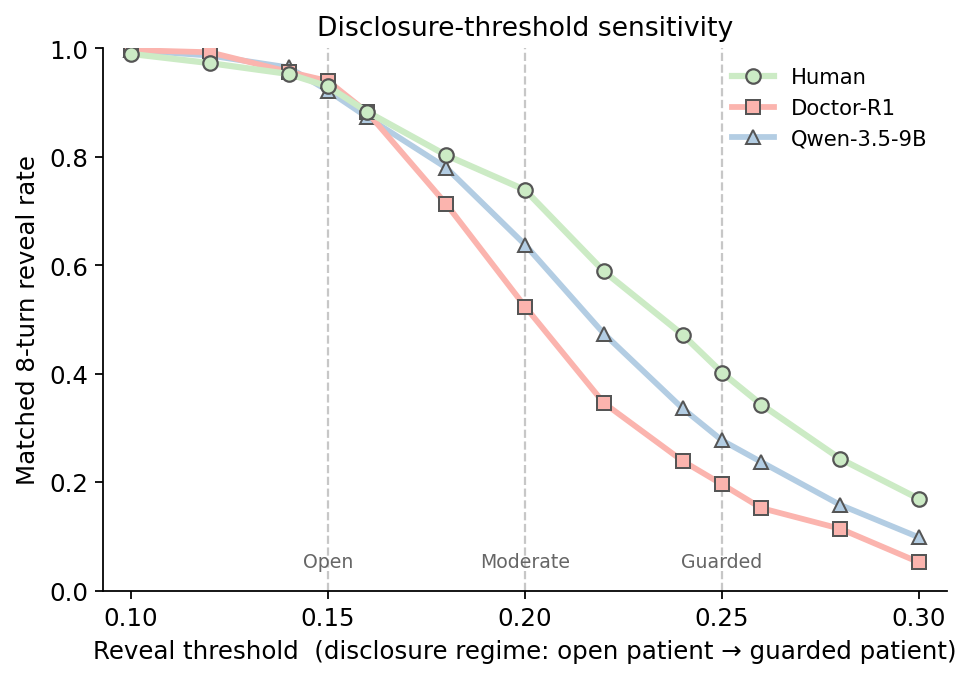}
\caption{Matched 8-turn reveal rate after recomputing the state trace over fixed transcripts across reveal thresholds. Lower thresholds represent more readily disclosing patients and higher thresholds more guarded patients. All systems decline as disclosure tightens, with the human trace most robust in the guarded regime. Because doctor and patient utterances are fixed in this diagnostic sweep, the absolute values are not directly comparable to the gold-matched headline Reveal Rate.}
\label{fig:disclosure_sensitivity}
\end{figure*}

The fixed-transcript sweep in Figure~\ref{fig:disclosure_sensitivity} isolates the threshold effect. We additionally reran the AI dialogues end-to-end at the guarded threshold ($0.25$): gold-matched Reveal Rate falls from 0.348 to 0.156 for Doctor-R1 and from 0.431 to 0.250 for Qwen. This confirms the same directional effect under closed-loop re-simulation.

\subsection{Performance by Concern Type and Severity}
\label{sec:appendix_concern_breakdowns}
\begin{figure*}[htbp]
\centering
\includegraphics[width=\textwidth]{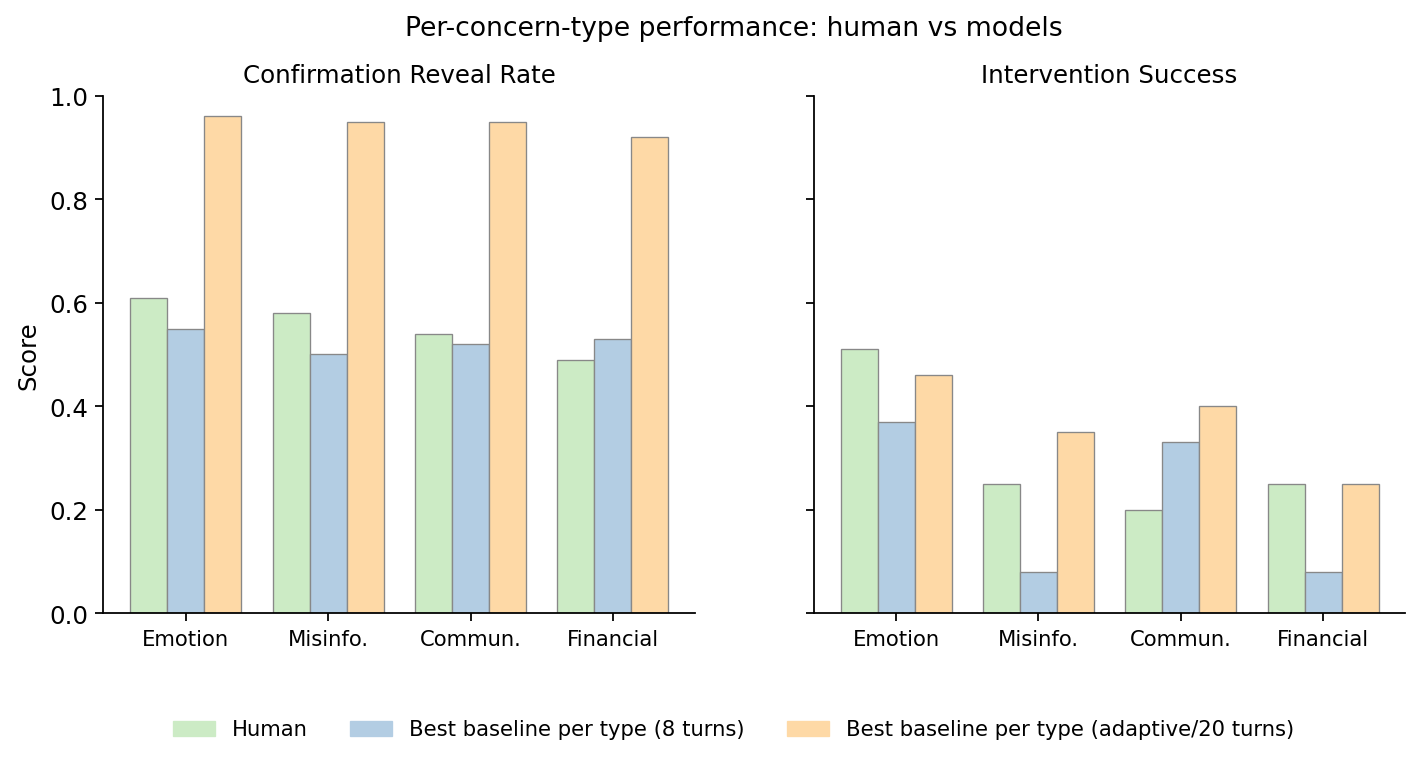}
\caption{Performance by hidden-concern type. ``Best baseline per type'' selects the strongest evaluated model separately for each type and setting; it is not one fixed model. Extended confirmation uses the adaptive policy, and extended intervention uses 20 turns.}
\label{fig:concern_type_performance}
\end{figure*}

Figure~\ref{fig:concern_type_performance} disaggregates the two task outcomes by the four concern types. At the matched 8-turn setting, the human reference exceeds the strongest model result on three of four types in each task. Extended runs raise confirmation Reveal Rate to 0.92--0.96 across types, but require approximately 19--20 turns and do not yield a proportional intervention improvement.

We also assigned each primary concern a low, moderate, or high severity label using a \texttt{gpt-5-mini} rubric adapted from distress and supportive-care-needs instruments \citep{roth1998rapid,boyes2009brief}. Severity denotes how strongly an unelicited and unaddressed concern could interfere with coping or care, not the biomedical acuity of the presenting condition.

{\small
\setlength{\tabcolsep}{5pt}
\renewcommand{\arraystretch}{1.08}
\begin{table}[htbp]
\centering
\begin{tabular}{lrrr}
\toprule
Severity ($n$) & Human conf. Reveal & Human interv. Success & Doctor-R1 Success \\
\midrule
Low (6) & 0.625 & 0.667 & 0.333 \\
Moderate (219) & 0.541 & 0.411 & 0.416 \\
High (75) & 0.601 & 0.453 & 0.467 \\
\bottomrule
\end{tabular}
\caption{Outcomes by primary-concern severity. Doctor-R1 uses the 20-turn intervention setting. The low-severity stratum is too small for interpretation.}
\label{tab:severity_breakdown}
\end{table}
}

The benchmark therefore concentrates on consequential concerns: 294/300 are moderate or high. The human--model relationship is similar in the two interpretable strata. A blind audit of 100 severity labels agrees exactly on 75, differs by at most one adjacent level on all 100, and has no low-versus-high disagreement (unweighted $\kappa=0.374$; quadratic-weighted $\kappa=0.448$). The residual uncertainty is concentrated at the subjective moderate/high boundary.

\subsection{Human-Reference Coverage and Dialogue Length}
\label{sec:appendix_human_robustness}
The clinician cohort is a real-clinician reference under matched observable information, not an expert-performance ceiling. It is relatively young, geographically skewed, and dominated by clinicians with 1--5 years of experience. We therefore stratified all 300 case-level outcomes by experience and first-language status. Early-career includes students/trainees and clinicians with at most five years of experience; all recruited participants met the English-fluency criterion.

{\scriptsize
\setlength{\tabcolsep}{3.6pt}
\renewcommand{\arraystretch}{1.08}
\begin{table*}[htbp]
\centering
\resizebox{\textwidth}{!}{%
\begin{tabular}{lccccc}
\toprule
Contrast & Interv. Success & Interv. Reveal & Conf. Reveal & Conf. Coarse F1 & Conf. Fine F1 \\
\midrule
Native / non-native English & 0.482 / 0.405 & 0.529 / 0.470 & 0.688 / 0.504* & 0.563 / 0.579 & 0.528 / 0.522 \\
Early-career / senior & 0.417 / 0.451 & 0.482 / 0.500 & 0.559 / 0.556 & 0.570 / 0.586 & 0.524 / 0.522 \\
\bottomrule
\end{tabular}
}
\caption{Human-reference subgroup outcomes. *The native/non-native confirmation Reveal contrast is significant ($p<0.001$); the other nine contrasts are non-significant ($p>0.2$ for scalar metrics or bootstrap 95\% intervals include zero for F1).}
\label{tab:human_subgroup_robustness}
\end{table*}
}

Experience does not materially change any reported outcome in this sample. First-language status changes confirmation Reveal Rate but not submitted-finding F1 or intervention Success, so the subgroup analysis supports the aggregate intervention reference while exposing a language-linked elicitation difference that deserves targeted study.

\begin{figure*}[htbp]
\centering
\includegraphics[width=0.78\textwidth]{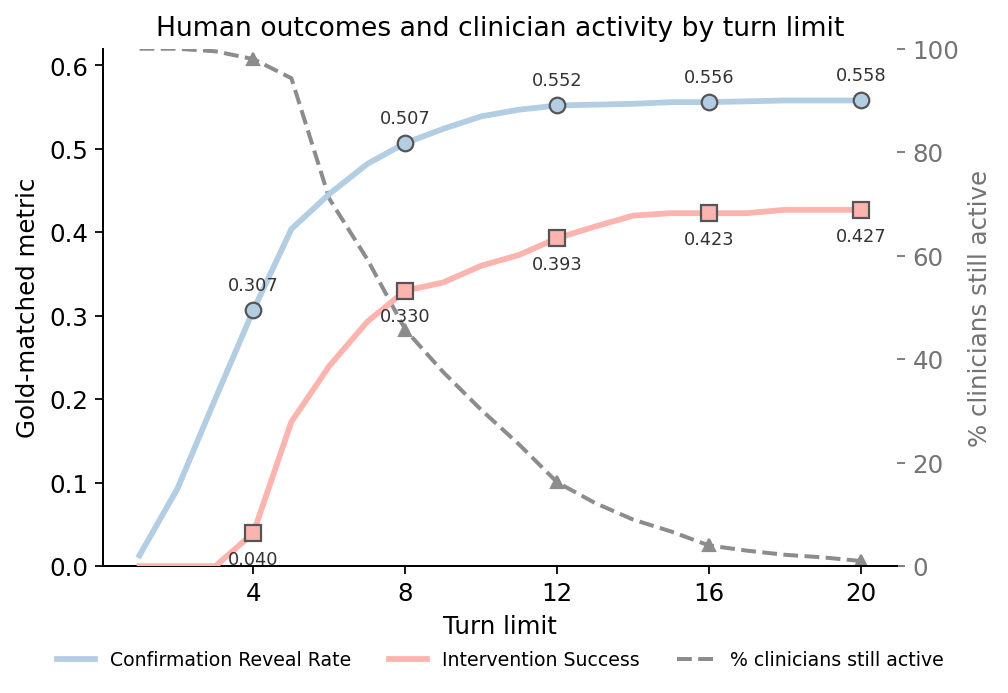}
\caption{Cumulative human-reference outcomes and the share of clinicians still active by turn limit. From turn 12 to 20, Reveal Rate increases from 0.552 to 0.558 and Success from 0.393 to 0.427, while the active share falls below one fifth after turn 12.}
\label{fig:human_turn_plateau}
\end{figure*}

Figure~\ref{fig:human_turn_plateau} shows that the observed human outcomes are close to their empirical plateau under the study protocol. This does not make a fixed 20-turn human condition equivalent to the collected condition, but it bounds how much of the headline comparison is attributable to the longest observed human dialogues.

\subsection{Scope, Safety, and External Validity}
\label{sec:appendix_scope_limitations}
\paragraph{Reconstructed concerns.}
The hidden concerns originate in patient-authored posts and follow-up text, are structured by annotators, and are reviewed for clinical plausibility; they are therefore source-evidenced rather than invented from clinician intuition. Anonymous online discussions are useful for this construct because concerns withheld in ordinary encounters may be stated in that setting \citep{levy2018prevalence,barry2000unvoiced,snell2025assessing}. Nevertheless, the benchmark does not prospectively confirm with the original patients that these were the concerns they intentionally withheld. The labels should be interpreted as patient-text-supported reserved concerns, not ground-truth mental states.

\paragraph{Representativeness and external validation.}
\texttt{r/AskDocs} posters are self-selected, digitally literate enough to describe their situation online, and predominantly represented through English-language text. The benchmark does not estimate population prevalence and underrepresents limited digital access, limited English proficiency, very low health literacy, urgent encounters, and disclosure styles that the current simulator cannot sustain reliably. Human participants knew that the patient was AI-simulated, and the shared simulator is a controlled fixture for human--model comparison rather than evidence that its behavior matches real patients in every respect. We did not include a human standardized-patient control. Prospective studies with trained standardized patients, patient-confirmed concerns, multilingual encounters, and experimentally controlled disclosure styles are the appropriate next external-validation step \citep{barrows1993overview,bosse2015cost}.

\paragraph{Single target plan.}
Intervention Success uses one source-derived target plan to isolate the communication problem from open-ended plan generation. This increases internal consistency but can penalize clinically reasonable alternatives. For example, redirecting a cost-constrained patient from an emergency department to an accepted urgent-care alternative can fail the exact target gate. The current result is therefore a single-target variant; multi-plan evaluation with shared decision-making is future work.

\paragraph{Safety screen.}
\benchmark is not a complete assessment of medical correctness, contraindications, or guideline concordance. As a complementary check, a \texttt{gpt-5-mini} judge screened 100 Doctor-R1 and 100 human intervention consultations using the same strict criterion: flag only clearly harmful, contraindicated, or medically inappropriate advice that was affirmatively stated. One Doctor-R1 consultation (1\%) and three human consultations (3\%) were flagged. This sampled LLM-judge screen indicates that Doctor-R1's communication success is not obviously obtained through more unsafe advice, but it is not clinician adjudication and is not powered to establish comparative safety.

\section{Implementation Details}
\subsection{Turn Evaluator and Latent Policy}
\label{sec:appendix_turn_evaluator}
Every clinician turn is first passed through a shared runtime turn evaluator. The evaluator sees the current clinician utterance, a short window of recent clinician turns, and any currently pending patient clarification question. It returns a structured JSON payload containing a discrete clinician intent label, intent probabilities, a pending-question coverage flag, empathy strength, and the 10 rubric features in Table~\ref{tab:rubric_features}. This shared evaluator is used for both human-trace export and offline AI evaluation so that reveal and address dynamics are computed from the same signal space.

The latent policy converts these turn-level rubric signals into state transitions. For confirmation, the policy computes a reveal-observation probability independently for each still-hidden concern using a learned logistic model over rubric features plus a concern-target overlap score between the clinician utterance and the concern text. Reveal evidence is then updated with an exponential moving average and passed through hysteresis: a concern unlocks either when evidence exceeds a high threshold or when it exceeds a lower threshold for a required number of consecutive turns. If progressive reveal mode is enabled, these thresholds can tighten as more concerns have already been surfaced. In the released benchmark configuration, meta-category probing is hard-blocked before any reveal update is credited.

For intervention, a separate logistic model maps the same rubric feature vector to an addressing observation probability. Address evidence is tracked with its own exponential moving average. A concern can move from revealed to addressed only after the minimum reveal-to-address lag has elapsed, the per-turn addressing observation passes the configured quality gate, and the addressing evidence exceeds the high threshold for the required number of consecutive eligible turns. This makes intervention harder to game with a single persuasive sentence and aligns success with sustained, concern-specific engagement rather than superficial plan repetition.

The current deployment uses a learned latent-policy configuration when offline-fitted weights are available. Those weights are fit from batched pseudo-labels produced by the offline turn-labeling pipeline, with separate base coefficients and concern-family-specific deltas for reveal and address models; candidate threshold settings are then replayed offline before the runtime policy is frozen for benchmark evaluation. When weights are unavailable, the implementation retains a judge-based fallback path for debugging and backward compatibility, but the benchmarked runs described in the draft use the shared evaluator plus latent-policy path as the primary transition mechanism.

\subsection{Rubric Feature Vector}
\label{sec:appendix_rubric_features}

The turn-level rubric feature vector combines adapted signals from three established clinical communication frameworks with several task-specific project-defined features. Data gathering, emotional responsiveness, and partnership/activation are adapted from RIAS \citep{roter2002roter}; concern elicitation and space provision are adapted from VR-CoDES \citep{del2017verona}; and necessity support together with concern mitigation are adapted from the Necessity--Concerns Framework \citep{horne2013understanding}. The remaining dimensions (plan specificity / safety-net, pending-question coverage, and meta-probe risk) are task-specific rubric features introduced for this benchmark.

{\scriptsize
\setlength{\tabcolsep}{3pt}
\renewcommand{\arraystretch}{1.12}
\begin{table}[htbp]
\centering
\resizebox{\columnwidth}{!}{%
\begin{tabular}{p{0.10\linewidth} p{0.25\linewidth} p{0.15\linewidth} p{0.5\linewidth}}
\toprule
Symbol & Name & Framework & Interpretation\\
\midrule
$r^{\text{DG}}_t$ & Data gathering & RIAS & How strongly the turn elicits medically relevant information (questions, clarifications). \\
$r^{\text{ER}}_t$ & Emotional responsiveness & RIAS & Acknowledgment/validation of patient affect and concerns. \\
$r^{\text{PA}}_t$ & Partnership/activation & RIAS & Collaborative stance (shared decision-making cues, inviting preferences). \\
$r^{\text{CE}}_t$ & Concern elicitation & VR-CoDES & Degree to which the turn invites disclosure of worries/concerns. \\
$r^{\text{SP}}_t$ & Space provision & VR-CoDES & Whether the turn leaves room to elaborate (non-cutting, open invitation). \\
$r^{\text{NS}}_t$ & Necessity support & NCF & Strength of rationale for why the recommended plan is needed/beneficial. \\
$r^{\text{CM}}_t$ & Concern mitigation & NCF & Extent the turn directly addresses barriers/concerns (e.g., risks, misconceptions). \\
$r^{\text{PS}}_t$ & Plan specificity / safety-net & Task rubric & Specific, actionable plan details and clear safety-netting (what to watch for, when to escalate). \\
$r^{\text{PQC}}_t$ & Pending-question coverage & Task rubric & Degree the clinician addresses the patient’s most recent clarifying question (if any). \\
$r^{\text{MR}}_t$ & Meta-probe risk & Task rubric & Risk of category/checklist probing (anti-gaming signal), e.g., asking in labels instead of natural dialogue. \\
\bottomrule
\end{tabular}
}
\caption{Turn-level rubric feature vector $z_t$ used as model inputs. Framework-linked dimensions are adapted from RIAS, VR-CoDES, and the Necessity--Concerns Framework; task-rubric dimensions are project-defined.}
\label{tab:rubric_features}
\end{table}
}

\section{Usage of LLMs}
\label{sec:appendix_ai_usage}
LLMs were used solely as general-purpose assistive tools for tasks such as improving writing clarity and suggesting code snippets. All research design, analysis, and substantive writing were conducted by the authors, who bear full responsibility for the content of this work.